\documentclass{article} %
\usepackage{iclr2027_conference,times}

\PassOptionsToPackage{unicode}{hyperref}
\PassOptionsToPackage{hyphens}{url}
\usepackage{xcolor}
\usepackage{amsmath,amssymb}

\usepackage{hyperref}
\usepackage{caption}
\definecolor{citeblue}{HTML}{1565C0}
\definecolor{linkred}{HTML}{B71C1C}
\definecolor{urlpurple}{HTML}{6A1B9A}
\usepackage{wrapfig}
\hypersetup{
colorlinks = true,     %
citecolor  = citeblue,
linkcolor  = linkred,
urlcolor   = urlpurple,
pdftitle   = {Open-world AI: a Tale of Three Requirements},
pdfauthor  = {Jianyu Zhang, L\'eon Bottou}, 
}

\usepackage{iftex}
\ifPDFTeX
  \usepackage[T1]{fontenc}
  \usepackage[utf8]{inputenc}
  \usepackage{textcomp} %
\else %
  \usepackage{unicode-math} %
  \defaultfontfeatures{Scale=MatchLowercase}
  \defaultfontfeatures[\rmfamily]{Ligatures=TeX,Scale=1}
\fi

\usepackage[most]{tcolorbox}
\usepackage{amsthm}

\newtcolorbox{coloredproof}{
  colback=gray!25,
  colframe=black,
  breakable,
  boxrule=0pt,
}
\usepackage{lmodern}
\IfFileExists{upquote.sty}{\usepackage{upquote}}{}
\IfFileExists{microtype.sty}{%
  \usepackage[]{microtype}
  \UseMicrotypeSet[protrusion]{basicmath}
}{}
\makeatletter
\@ifundefined{KOMAClassName}{%
  \IfFileExists{parskip.sty}{%
    \usepackage{parskip}
  }{%
    \setlength{\parindent}{0pt}
    \setlength{\parskip}{6pt plus 2pt minus 1pt}}
}{%
  \KOMAoptions{parskip=half}}
\makeatother
\usepackage{enumitem}
\usepackage{longtable,booktabs,array}

\usepackage{calc}
\usepackage{etoolbox}
\usepackage{pifont}
\usepackage{float}
  \usepackage{colortbl}          %
  \definecolor{diagred}{RGB}{253,224,224}
  \definecolor{longblue}{RGB}{222,235,250}
  
\makeatletter
\patchcmd\longtable{\par}{\if@noskipsec\mbox{}\fi\par}{}{}
\makeatother
\IfFileExists{footnotehyper.sty}{\usepackage{footnotehyper}}{\usepackage{footnote}}
\makesavenoteenv{longtable}
\usepackage{graphicx}
\usepackage{bookmark}
\IfFileExists{xurl.sty}{\usepackage{xurl}}{}
\usepackage[disable]{todonotes}

\title{Open-world AI: a tale of three Requirements}

\author{Jianyu Zhang \\
FAIR, Meta\thanks{This research, including all access and use of models and datasets, was completed in author's personal capacity.} \\ %
\texttt{jianyu@nyu.edu} \\
\And
L\'eon Bottou \\
Flatiron Institute \\
\texttt{leon@bottou.org} \\
}

\newcommand*{\iid}{\textsc{iid}}

\newcommand*{\revise}[1]{{{{#1}}}}

\definecolor{main}{HTML}{5989cf}
\definecolor{sub}{HTML}{cde4ff}

\newtcolorbox{boxA}{
    boxrule = 1.pt,
    colframe = black,
    top         = 1pt,
    bottom      = 1pt, 
}

\newtcolorbox{boxH}{
    colback = sub,
    colframe = main,
    boxrule = 0pt,
    leftrule = 6pt,
    top         = 2pt,
    bottom      = 2pt, 
}
\newif\ifshowmarkup
\showmarkuptrue
\ifshowmarkup
  \newcommand{\NEW}[1]{{#1}}    %
\else
  \newcommand{\NEW}[1]{#1}
  
\fi
\iclrfinalcopy
\begin{document}
\maketitle
\begin{abstract}

The closed world asks AI for specialized competence and lets it manufacture the \iid~data its theory assumes: a fixed criterion licenses collecting examples until the task is covered. The open world asks for versatility --- taking up tasks a machine was not built for, from few examples and with little prior from a human designer --- and offers data that is grouped, not shuffled.
Its tasks are not known in advance, so a machine must do \emph{inference-time} \emph{learning} (1st requirement). Modeling tasks as sharing latent factors and decomposing the training objective, we derive two more: a \emph{rich representation}, retaining more than \iid~generalization requires, and a \emph{predictive disentanglement} that aligns inputs and outputs into matched low-dimensional groups.
The analysis further predicts the need for a consistent inner learner, and the benefit of long contexts spanning related tasks.
We test these claims on language models up to $8$\textsc{b}\footnote{Code: \url{https://github.com/facebookresearch/MemoryMosaics}}, asking whether a model has a property (qualitative) rather than how much it wins (quantitative). Models that meet the requirements improve as data from related tasks accumulates; the rest degrade, and neither scale nor layer-type interleaving repairs this.

\end{abstract}

\vspace{-0.5em}
\section{Introduction}

\begin{boxH}
\textit{We propose that a 2 month, 10 man study of artificial intelligence be carried out
during the summer of 1956 [...] An attempt will be made to find how to make machines [...]
solve kinds of problems now reserved for humans [...]. We think that a significant advance
can be made [...] for a summer.} \quad \quad ~~~--- McCarthy et al., 1955
\end{boxH}

Seventy years after the Dartmouth proposal \citep{mccarthy1955proposal},
artificial intelligence has produced a long sequence of successes associated with the
\emph{closed world}: artificial environments
\citep{samuel1959some,tesauro1995temporal,silver2016mastering} and
specific real-world tasks
\citep{matan1991multi,taigman2014deepface}.
Each rests on a clear criterion of success, which does two things at once. 1) The criterion specifies the
distribution the machine is measured against, licensing the collection of more iid examples, with a generalization gap falling as $O(\smash[t]{l^{-1/2}})$ in the number of examples $l$
\citep{vapnik1991principles}. 2) It reveals not merely \emph{whether} a
machine fails but \emph{how} it fails, so that designers can repair the problems one after another. Together these give a loop
along which capability accumulates: locate a failure, collect data against it, repair, repeat. On tasks other than the one it was trained on --- even one that looks close --- the machine may fail
\citep{zech2018variable,koh2021wilds}, and the loop must be run again, by a human, for each
new task. After building AI this way, rather than proving
that our machines are intelligent, we have more often found satisfaction in proving that we are.

Do large language models escape the \emph{closed world}? Their competence across many specialized domains is real and impressive, and the in-context learning ability it comes with is genuinely remarkable. 
But two questions are still worth asking: how that competence was
obtained, and what kind of in-context learning it is. 
On the first, the training practice is eager to consume all available data and to annotate as much task-specific data as possible until the tasks of interest are likely to be covered (\citealp[\S4.1]{yang2025qwen3}; \citealp[\S5.1]{deepseek2026v41flash}).\footnote{Web crawls, source code, book corpora, chain-of-thought for mathematics, queries for reasoning. The practice is general; these two reports document it explicitly, which most do not.}
Benchmarks then serve as both criterion and measurement; a failure
identified is a dataset collected, and the loop continues. This is the closed-world loop, \emph{industrialized} rather than escaped. 
On the second, modern in-context learning is impressive but fragile \citep{li2024long}: supplying \emph{more} demonstrations frequently makes performance \emph{worse} rather than better; and frontier models (e.g., \textsc{gpt}-4o) may under-perform smaller \& older ones (e.g., \textsc{llama}-2-7b). Neither
more task-specific \iid~data nor more scale reliably improves performance on the task at hand.
This is not yet learning.

The Dartmouth proposal set out to produce versatility rather than specialized competence --- a machine that can take up a task it was not built for. We take that as the target:
\vspace{-0.1em}
\begin{boxA}
AI for the open world requires a machine to learn a \textbf{wide range} of new tasks, using
\textbf{fewer examples} and \textbf{less task-specific prior knowledge} from human designers.
\end{boxA}
Unlike closed-world AI where the \emph{tasks of interest} were already known at training time, open-world AI does not come with predefined lists of tasks. Thus,  models must be able to adapt to new tasks on the fly.\footnote{We consider this well understood and broadly accepted.} This gives the first requirement, \emph{inference-time learning}.

\begin{figure}
    \vspace{-1.5em}
    \centering
    \includegraphics[width=0.75\linewidth]{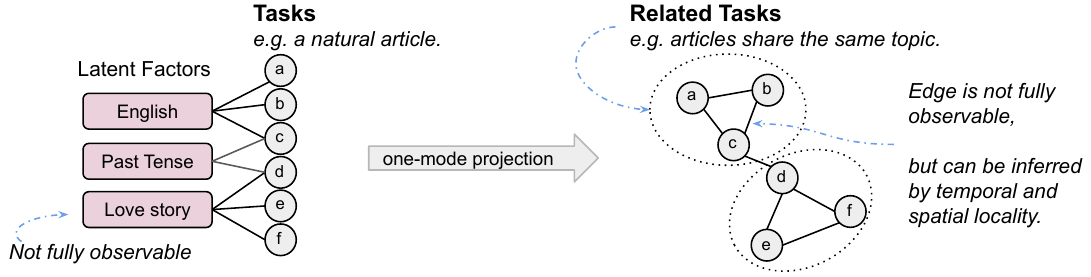}
    \caption{Latent factors, task, and related tasks. }
    \label{fig:related-tasks}
    \vspace{-1.5em}
\end{figure}

Just as closed-world AI relies on the \iid~sampling assumption to connect training and test \emph{examples}, open-world AI needs connections between training and test \emph{tasks} (groups of examples) \citep{baxter2000model,Maurer2016,tripuraneni2020theory,du2020few}.
Fortunately, the world is structured, and tasks are not arbitrary. For example, in language modeling, poetry and mathematics are different tasks following very different statistics. Yet these two tasks share substantial latent factors, e.g., language and grammatical construction. This observation suggests viewing tasks and latent factors as a bipartite graph (as in Figure~\ref{fig:related-tasks}), and assuming two tasks to be more similar when they share more latent factors.\footnote{Without this
    assumption, the world would be arbitrary and unpredictable.}  

In this work, Section \ref{sec:inference-time-learning}, we first analyze \emph{inference-time learning} under the assumption that the latent-factor overlap of \emph{similar tasks} can be expressed as an overlap over internal representation of either the model inputs or the model outputs. We argue that this analysis suggests two additional requirements: 

\begin{itemize}[leftmargin=0.8cm]
    \setlength\itemsep{-0.4em}
    \item \emph{rich representation} (gather broadly): learns features richer than
    \iid~generalization requires, making a {wide range} of new tasks reachable;
    \item \emph{predictive disentanglement} (group predictively): disentangles input and output representations into matching groups, and performs prediction within each group, making a new task learnable from {fewer examples}.
\end{itemize}

Furthermore, this analysis allows us to make interesting predictions about learning algorithms for building open-world AI: 
\begin{itemize}[leftmargin=0.8cm]
    \setlength\itemsep{-0.4em}
    \item the importance of a consistent inner learner for inference-time learning;
    \item the benefit of long contexts that span multiple related tasks, provided the target and related tasks are handled separately.
\end{itemize}

Finally, Section~\ref{sec:exps} tests these requirements and predictions on language models up to $8$\textsc{b}.
What we look for is a \emph{qualitative} difference --- whether a model has a given property or
lacks it --- rather than a \emph{quantitative} margin.
A margin in today's language modeling depends sensitively on engineering detail that is largely
unshared and would not fit in a scientific paper in any case; a qualitative difference is robust
to that detail.

What we find is that models meeting the requirements improve as data from related tasks
accumulates, while the rest degrade. Neither scale nor layer-type interleaving repairs it.

\section{Inference-time learning}
\label{sec:inference-time-learning}
In closed-world AI, both training and test examples are assumed to be drawn \iid. There is then no need to learn at inference time, provided the data are actually sampled that way. 
Real-world data, however, has been observed for half a century to be ``grouped'', where a group of examples (a \emph{task}, written $\tau$) is sampled conditionally on the choice of certain latent factors \citep{heckman1979sample, zadrozny2004learning, Torralba11, koh2021wilds, liu2025decade}. The latent factors are not fully observable and sparsely changed \citep{scholkopf2021toward} over tasks, resulting in a series of similar but not identical tasks at inference time. Thus, in open-world AI, adapting to new tasks on the fly --- \emph{inference-time learning} --- is necessary. 

\subsection{Modeling inference-time learning}
\label{sec:principles}

To model inference-time learning, we make two assumptions: 1) inputs and outputs both have internal representations that are shared across tasks \citep{finn2017model, garnelo2018conditional,kim2019attentive}; 2) the connection between those representations is task-specific, and must therefore be learned at inference time.
The first says what is carried over between tasks; the second says what is not. Together, they ask a machine to have four parts:
\begin{enumerate}[leftmargin=0.8cm]
\setlength\itemsep{-0.4em}
    \item[\textcolor{blue}{\ding{202}}] \textbf{input feature extractor} $\phi_\theta,$ on input $x$, $ z = [z_1,\ldots,z_G] = \phi_\theta(x)$, shared across tasks.
  \item[\textcolor{blue}{\ding{203}}] \textbf{output feature extractor} $\psi$, on output $y$, $\gamma=[\gamma_1,\ldots,\gamma_G] = \psi(y)$, shared across tasks. $\phi$ and $\psi$ learn representation and partition them into matching groups $\{(z_g, \gamma_g)\}$. 
\item[\textcolor{blue}{\ding{204}}] \textbf{decoder} (affine) $A$ maps $\gamma$ back to $y$: $y = A\gamma$.\footnote{An affine $A$ does not make $\psi$ affine, since $\gamma$ lives in a larger space than $y$.}
    \todo{affine decoder.}
  \item[\textcolor{blue}{\ding{205}}] \textbf{inner learner} $f$, the only task-specific part, \emph{adapted at inference time}. It is fitted on the $l$ support-set examples $ S^{(l)}=\{(x_i,y_i)\}_{i=1}^l\sim\tau$, carried into latent space as $\{(z_g,\gamma_g)\}$, and predicts group-wise on a query $x^{*}\sim\tau$\,:
  \(
   \hat\gamma^{}_g = f_g(z^{}_g)\,,
  \hat y^{} = A[\hat\gamma^{}_1,\ldots,\hat\gamma^{}_G],
\)
where each $f_g$ is fitted on the group-$g$ pairs $\{(z_g,\gamma_g)\}$ alone.
\end{enumerate}

In practice a model stacks many such units; the analysis concerns the functional decomposition, not its allocation across layers.

\paragraph{\textit{Training objective}}
The training process minimizes the error (for ease of analysis we
take the squared error throughout) on a query \((x^*, y^*)\) given support-set \(S=\{(x,y)\}\) drawn from the same task, averaged over tasks: 
\begin{equation}\label{eq:obj}
\mathbb{E}_{\tau, S, (x^*,y^*)}\bigl[(\hat{y}^* - y^*)^2\bigr]
\;=\; \mathbb{E}_{l\sim p(l)}\;
      \mathbb{E}_{\tau, S^{(l)}, (x^*,y^*)}\bigl[(\hat{y}^* - y^*)^2\bigr] ,
\end{equation}
\NEW{where $p(l)$ is the distribution of support-set sizes, which deployment does not let
a machine choose.}

\subsection{Decomposing the training objective}
\label{sec:decompose}
The reconstruction of $y$ directly supervises decoder and output extractor to avoid collapse: if $\gamma$ has dropped information about $y$, and equally if $A$ cannot read it back, the reconstruction fails. We therefore leave the decoder $A$ out of the analysis below. 
\todo{handle affine A. then ignore the analysis of A. }   
We decompose the training objective around the \emph{Bayes-optimal prediction} on query $x^*$ given the support set $S^{(l)}$, formed in the latent space that $\phi_\theta$ and $\psi$ provide:
\begin{equation}\label{eq:bayes}
\bar y^{*} \;=\; A\;\mathbb{E}\bigl[\gamma^{*} \mid z^{*}, \{(z,\gamma)\}^{(l)}\bigr].
\end{equation}
It is the best any inner learner could do with what the representations hand it. Since $A$ is affine it commutes with the conditional expectation, and since $A$ recovers $y$ from $\gamma$, conditioning on $\{(z,\gamma)\}^{(l)}$ is conditioning on $\{(z,y)\}^{(l)}$; hence $\bar y^{*} = \mathbb{E}[y^{*}\mid z^{*}, \{(z,y)\}^{(l)}]$. Predicting in latent space and predicting in output space share the same reference, and the decomposition can be written on $y$:
\begin{align}
\label{eq:decompose}
\mathbb{E}_{\tau, S^{(l)}, (x^*,y^*)}\bigl[(\hat{y}^* - y^*)^2\bigr]
&= \mathbb{E}_{\tau, S^{(l)}, (x^*,y^*)}
   \bigl[(\hat{y}^* - \bar{y}^* + \bar{y}^* - y^*)^2\bigr] \notag \\
&= \mathbb{E}_{\tau}\,
   \underbrace{\mathbb{E}_{S^{(l)}, x^*}\bigl[(\bar{y}^* - y^*)^2\bigr]}
   _{\ell_{\theta}(l)}
 \;+\; \mathbb{E}_{\tau}\,
   \underbrace{\mathbb{E}_{S^{(l)}, x^*}\bigl[(\hat{y}^* - \bar{y}^*)^2\bigr]}
   _{\ell_{f}(\theta, l)} \notag \\
&= \mathbb{E}_\tau \sigma_\tau^2 + \mathbb{E}_{\tau}[\ell_\theta(l) - \sigma_\tau^2] + \mathbb{E}_\tau \ell_f(\theta,l)\,,
\end{align}
where \(\sigma^2_\tau := \mathbb{E}_{x^*}\bigl[\mathrm{Var}(y^*\mid x^*,\tau)\bigr]\) is the irreducible error of task \(\tau\). The cross term $\mathbb{E}_{\tau, S^{(l)}, x^*}[(\hat y^{*} - \bar y^{*})(\bar y^{*} - y^{*})] = 0$ vanishes by the definition of $\bar y^{*}$: $\hat y^{*} - \bar y^{*}$ is a function of $(z^{*}, S^{(l)})$, and $\bar y^{*} - y^{*}$ has zero mean given the same.
This orthogonality is not special to squared error: it holds for any Bregman divergence, hence for cross-entropy.

Neither term in Eq.~\eqref{eq:decompose} can be optimized on its own: \(\ell_f\) is defined against the Bayes-optimal \(\bar y^{*}\), which is unknown. Only the full objective is available, and \(l\) decides which term it reaches. At large \(l\), \(\ell_f\) vanishes for any consistent inner learner, so reducing the objective reduces \(\ell_\theta\) --- the information the representation has permanently discarded --- and builds a \emph{rich representation}. At small \(l\), \(\ell_f\) dominates instead, and reducing the objective presses \(\phi\) and \(\psi\) toward coordinates in which the prediction \(z_g\rightarrow\gamma_g\) factorizes group-wise: \emph{predictive disentanglement}. The two are ordered: because \(\bar y^{*}\) is defined relative to \(\phi\) and \(\psi\), the small-\(l\) pressure is worth applying only once the representation is rich\footnote{This matches the finding of \citet{zhang2022rich, chen2023towards}, where algorithms that seek an invariant representation directly fail, and a rich representation restores them.}  (built at large \(l\)).

\paragraph{\uppercase\expandafter{\romannumeral1}.~~~\(\mathbb{E}_{\tau}[\ell_\theta(l) - \sigma_\tau^2]\) at large \(l\): rich representation.}

Bayes-optimal prediction \(\bar{y}^*\) uses the exact prior of each task (ideal case), and thus depends only on the information retained in $\phi_\theta$, not on how it is arranged. \(\mathbb{E}_{\tau}[\ell_\theta(l) - \sigma_\tau^2]\) removes the effect of tasks' irreducible error; at large $l$ it measures the information about $p(y\mid x,\tau)$ that $\phi_\theta$ has permanently discarded. A smaller \(\mathbb{E}_{\tau}[\ell_\theta(l) - \sigma_\tau^2]\) indicates a richer representation \citep{zhang2023learning}, which may be redundant for one task but helpful for a broader range of tasks.

Furthermore, this work provides a way to learn rich representation. Call an inner learner \emph{consistent} if $\ell_f(\theta,l)\to0$ as $l\to\infty$ --- given enough examples from a task, its prediction approaches the Bayes-optimal $\bar y^{*}$ \citep{gyorfi2002distribution}.\footnote{This excludes learners whose effective memory is limited, whether by a fixed-size state (as in standard recurrent and state-space models), a sliding window, or strongly decaying positional encodings.} With a consistent inner learner $\mathbb{E}_\tau\ell_f$ vanishes at large $l$, so the reduction of \(\mathbb{E}_{\tau, S^{(l)}, (x^*,y^*)}\bigl[(\hat{y}^* - y^*)^2\bigr]\) operates directly on \(\mathbb{E}_{\tau}[\ell_\theta(l) - \sigma_\tau^2]\).
\emph{It implies that a rich representation can be constructed by minimizing \(\mathbb{E}_{\tau, S^{(l)}, (x^*,y^*)}\bigl[(\hat{y}^* - y^*)^2\bigr]\) at large $l$.} 
A linear Gaussian case analysis on rich representation is shown in Appendix~\ref{apx:linear_gaussian}.

\paragraph{\uppercase\expandafter{\romannumeral2}.~~~\(\ell_f\) at small \(l\): predictive disentanglement.}
The term \(\ell_f\) measures how far the actual inner learner is from the Bayes-optimal predictor (which uses the exact task prior) under the learned representation.
Once the representation is rich, \(\ell_f\) dominates in the small-\(l\) regime. So the goal is no longer to refine the representation, but to let the inner learner reach \(\bar{y}^*\) from as few examples as possible.
Without being given a task-specific prior, this asks for a prior shared across the tasks of interest. One way to supply it is to reduce the inner learner from a single joint prediction $z\rightarrow\gamma$ with $z,\gamma \in \mathbb{R}^m$, to $G$ separate predictions $z_g\rightarrow \gamma_g$ with $z_g, \gamma_g \in \mathbb{R}^{m/G}$.\footnote{In analogy to the sparse mechanism shift hypothesis \citep{scholkopf2021toward}: tasks vary along sparse, independent directions.} Feature groups are shared across tasks; tasks differ in the function mapping within each group:
\begin{align}
\label{eq:partition}
 z=\phi(x)=(z_1,\ldots,z_G),\qquad \gamma=\psi(y)=(\gamma_1,\ldots,\gamma_G),  \notag \\
\mathbb{E}[\gamma\mid z,\tau] =
    \bigl(
        \eta_{\tau 1}(z_1),\ldots,\eta_{\tau G}(z_G)
    \bigr),
    \qquad
    \eta_{\tau g}(z_g):=\mathbb{E}[\gamma_g\mid z_g,\tau].
\end{align}
Under this structure the Bayes predictor itself factorizes group-wise, so a \emph{block}-diagonal inner learner loses nothing. Therefore, minimizing \(\ell_f\) at small \(l\) with such a learner pressures the representations \(\phi,\psi\) to find the coordinates in which this form holds. Since $(z_g, \gamma_g)$ are aligned for \emph{prediction}, $(z_1,\dots,z_G)$ and $(\gamma_1, \dots, \gamma_G)$ are \emph{disentangled} in groups; we call this \emph{predictive disentanglement}.

\paragraph{\textit{Prediction in latent space.}}
 Predicting $\gamma=\psi(y)$ rather than $y$ is the move made explicit by joint-embedding predictive architectures (\textsc{jepa}) \citep{lecun2022path}\footnote{Contrastive predictive coding \citep{oord2018representation} makes the same move with a contrastive objective. Joint-embedding methods (\textsc{jea}) \citep{chen2020simple,bardes2022vicreg} match the embeddings of two views rather than predicting one from the other; they have no predictor for Eq.~\eqref{eq:partition} to factorize.}, which encode the output and measure prediction error in embedding space to avoid spending capacity on unpredictable detail. \emph{Predictive disentanglement} shares that step but differs in two respects. First, the predictor is not dense, but group-wise as in Eq.~\eqref{eq:partition}. Its sample complexity is governed by the sub-problem dimension $m/G$ rather than the full dimension $m$ \citep{stone1985additive}. Second, the motivation of predictive disentanglement is sample complexity at small $l$ rather than the difficulty of the output space. \textsc{jepa} is recovered with a single dense group, no task conditioning, and no decoder.

\subsection{Two predictions}
\label{sec:two_predictions}
\subsubsection{Consistency of the inner learner}
The rich-representation analysis depends on the consistency of the inner learner: a consistent inner learner drives \(\ell_f\) to zero at large \(l\), so the optimization pressure falls on \(\phi_\theta\) and enriches it. This rich representation in turn prepares features for learning predictive disentanglement at small \(l\).

\subsubsection{Task size, and the use of related tasks}
\label{sec:lambda}
The analysis of rich representation starts from large \(l\). Without large $l$, a rich representation cannot be constructed, and predictive disentanglement is therefore hard to learn. The training process thus requires each task to contain enough examples.

Could we sample arbitrarily many examples from the target task, the inner learner would need
only to be consistent --- predictive disentanglement having already made the sample complexity
manageable --- and would handle new tasks at inference time without a designer's task-specific
prior. We cannot. As the ImageNetv2--ImageNet comparison shows \citep{recht2019imagenet},
unknown hidden factors shift the distribution even under a faithful replication protocol, so
data from the \emph{target} task \(\tau_t\) is hard to accumulate. What accumulates instead is data from
\emph{related} tasks \(\tau_r\), and using them when the target task does not supply enough examples
becomes essential.

We now analyze that case. In the feature space defined by Eq.~\eqref{eq:partition}, a related
task helps with the target task's group \(g\) when its group-wise conditional mean is close to the
target task's, \(\eta_{\tau_r g}(z_g):=\mathbb{E}[\gamma_g\mid z_g,\tau_r]\approx\eta_{\tau_t
g}(z_g)\) (dropping \(g\), \(z\) and \(\tau\) when unambiguous: \(\eta_r\approx\eta_t\)), and
hurts it otherwise. Pooling every example into one estimate cannot tell the two cases apart, and the damage grows as related tasks accumulate. Two things are therefore required: estimating the target and related tasks
\emph{separately}, as \(\hat\eta_t\) and \(\hat\eta_r\), and mixing them \emph{dynamically},
\[
    \hat\eta
    =
    \alpha\,\hat\eta_{t}
    +
    \bigl(1-\alpha\bigr)\hat\eta_{r}\,,
\]
where \(\hat\eta_{t}\) uses target examples alone, \(\hat\eta_{r}\) uses related-task examples, and \(\alpha\) indicates how to mix. The optimal \(\alpha\) grows as related tasks move further from the target task, and the learner effectively stops borrowing once they become too distant \citep{grant2018recasting}. 

\begin{coloredproof}
\begin{proof}
Write \(\hat\eta_{t}=\eta_{t}+\xi_{t}\) and \(\hat\eta_{r}=\eta_{r}+\xi_{r}\) with independent
zero-mean estimation errors, \(\mathrm{Var}_{t}:=\mathbb{E}[\xi_{t}^2]\), \(\mathrm{Var}_{r}:=\mathbb{E}[\xi_{r}^2]\),
and \(\delta:=\eta_{r}-\eta_{t}\) the mechanism discrepancy between them (\(\eta_{r}\) being the
pooled mean over related tasks when several are available). Minimizing the squared error
\(\mathbb{E}\bigl[(\hat\eta-\eta_{t})^2\bigr] = \alpha^2 \mathrm{Var}_{t}+(1-\alpha)^2(\mathrm{Var}_{r}+\delta^2)\)
gives the optimal weight:
\[
    \alpha^{*}
    =
    ({\mathrm{Var}_{r}+\delta^{2}})/({\mathrm{Var}_{t}+\mathrm{Var}_{r}+\delta^{2}})\,,
\]
which depends only on the statistics of the target and related task estimates.
\(\delta\) is not observed at inference. Treating tasks within a neighborhood as exchangeable makes \(\mathbb{E}[\delta^{2}]\) the between-task variance, and spread among the related tasks estimates it \citep{gelman2007data}.
\end{proof}
\end{coloredproof}

\section{Experiments}
\label{sec:exps}

\paragraph{Qualitative effects, not quantitative ones.}
Sections~\ref{sec:decompose} and \ref{sec:two_predictions} derive three requirements and two predictions. The experiments in this section ask, for each, whether a model has it or lacks it --- a \emph{qualitative} comparison (\textit{``does a model have a property needed for open-world AI?''}) rather than a \emph{quantitative} one (\textit{``the proposed model outperforms baselines by $x$ percent.''}).
A quantitative margin in modern language modeling depends sensitively on engineering details --- data curation, training schedules, evaluation schemes --- that are largely unpublished, making the comparison fragile. By contrast, a qualitative comparison asks for the sign of an effect rather than its size, and a sign is robust to those details.

\subsection{Connecting the experiments to the theory}
\label{sec:exp:setup}
 
\paragraph{Language data.}
The theory is stated over examples and tasks and does not depend on modality. We instantiate it
in language because language corpora are large, widely shared, and the architectures we compare
are established in that setting. Every model in this work is trained on exactly the same
mixture: Cosmopedia \citep{benallal2024cosmopedia}, NemotronCC-high \citep{su2025nemotron},
open-web-math \& arxiv \citep{azerbayev2024llemma}, Gutenberg books
\citep{manu2023gutenberg}, and ShareGPT \citep{sharegpt_vicuna_unfiltered}.\footnote{Topical diversity is what matters, not the choice of sources: the same experiments on Cosmopedia alone --- a single synthetic corpus, but a topically diverse one --- give the same conclusions.}

\paragraph{Building tasks and related tasks.} Following the illustration of \emph{tasks} and \emph{related tasks} in Figure~\ref{fig:related-tasks}, we use natural documents as \emph{tasks}, and semantic clusters of documents as \emph{related tasks}. We apply $k$-means clustering to document embeddings \citep{nussbaum2024nomic}, getting $1150$ clusters for training (each contains related tasks)\footnote{The exact number of clusters does not affect the conclusions; anything from a few hundred to a few thousand works.}. Similarly, we partition the Simple English Wikipedia (SimpleWiki)\footnote{\url{https://simple.wikipedia.org/wiki/Simple_English_Wikipedia}} evaluation corpus into 19 clusters, after removing template-stub clusters (Appendix~\ref{apx:eval_clusters}).

The clusters built this way are semantically meaningful and reflect real-world groupings.
To rule out artifacts of the $k$-means clustering, we also run the same evaluation without any post-clustering, treating Simple English Wikipedia as a single natural cluster. See Appendix~\ref{apx:simplewiki_no_cluster} for details.

\paragraph{Training and evaluation.}

Unless stated otherwise, every model at each scale uses the same hyper-parameters, optimizer, and amount of training data, with no per-model tuning. Each sequence is built by sampling documents (tasks) from one randomly drawn cluster (related tasks), delimiting each with \texttt{[bos]} and \texttt{[eos]}\footnote{Every model can therefore tell the target task apart from the related tasks in its context.}, concatenating them, and truncating at length $l$ for training, and at length $l^\prime$ for evaluation. The training objective, as in Eq.~\eqref{eq:obj}, averages the loss from position $1$ to $l$, \revise{so $l$ is the maximum support set size seen in training; we call it the \emph{maximum training length}, or \emph{training length} for short.}

\paragraph{How to read the evaluation.}
To keep the comparison \emph{qualitative}, and so robust to engineering choices and to the environment, we never directly compare raw performance numbers across model families. Instead, for each model or model family, we vary the other factors --- training length $l$, evaluation length $l^\prime$, how to use related tasks --- and ask whether the model has a property needed for open-world AI, yes or no.

\subsection{Models to compare}
\label{sec:exp:models}

\begin{figure}[H]
    \centering
    \includegraphics[width=\linewidth]{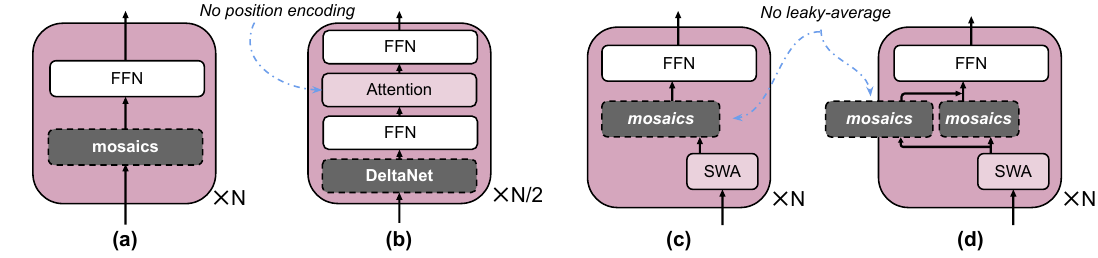}
        \caption{Model architecture groups. Architectures within each group differ only in the dark box. \textbf{(a)} \emph{Basic}: the dark box is one of $\{$Mosaics, attention, DeltaNet$\}$. \textbf{(b)} \emph{Modern hybrid}: $\{$DeltaNet, sliding window attention (\textsc{swa})$\}$, interleaved with full attention (\textsc{nope}). \textbf{(c)} \emph{$+$}: $\{$Mosaics, attention$\}$; the dark box skips the $s$ most recent positions, the \textsc{swa} below it uses a small window $w$, and the leaky average in Mosaics is removed. \textbf{(d)} \emph{Two-stage}: $\{$Mosaics, attention$\}$; as (c), but the target
    task and the related tasks are handled separately.}
    \label{fig:architecture}
\end{figure}

\paragraph{Architecture.}
\revise{We use two scales, $\sim$$1.5$B and $\sim$$8$B.
Every model has the same architecture layout: a token embedding, a stack of blocks each pairing a \emph{learning module} (the ``dark box'' of Figure~\ref{fig:architecture}) with a \textsc{s}wi\textsc{glu} feed-forward network, and a linear decoder.  A learning module contains key and value extractors (and, except in the Memory Mosaics family, a query extractor)\footnote{\label{fn:params}The Memory Mosaics family computes keys and queries with a single shared projection, so its learning module has 25\% fewer parameters ($3d^2$ instead of $4d^2$) than its counterparts.} and an inner learner that predicts from them. It runs $G$ heads in parallel, one per group $(z_g, \gamma_g)$ of Eq.~\eqref{eq:partition}: each head has its own extractors and its own group-wise inner learner $f_g$. Other detailed settings are in Appendix~\ref{apx:experiment_details}.}

\begin{itemize}[leftmargin=0.4cm]
\setlength\itemsep{-0.4em}
    \item[\textbf{a)}] \textbf{Basic architecture}. 
    \begin{itemize}[leftmargin=0.4cm]
    \setlength\itemsep{-0.4em}
      \item \emph{Memory Mosaics}: The learning module's inner learner is Gaussian kernel regression over keys and
  values, with the query equal to the key --- per head, a single \emph{long memory} spanning the whole context.
  This inner learner is consistent \citep{gyorfi2002distribution}.

 Keys $k_t$ and values $v_t$ are built from linear projections $W_k x_t$ and $W_v x_t$ of the input $x_t$. The
  key is a leaky average of past projections, $\bar k_t = g_t W_k x_t + \lambda_t \bar k_{t-1}$, $k_t = \bar k_t
  / \|\bar k_t\|$, with input-dependent $\lambda_t \in (0,1)$ and $g_t > 0$ \citep{zhang2026memory}; the value $v_t$ is a weighted average of the projections at $t$ and $t+1$ ($\|v_t\|=1$) \citep{zhang-2025}. Keys summarize the past and values the near future, so each stored pair is a \emph{predictive} association rather than a retrieval of one position's own representation. 
  The kernel regression predicts:
  \[\textstyle
      \frac{1}{Z_t}\sum_{i=1}^{t-1} e^{\,\beta\,k_t^{\top}k_i}\,v_i\,,
      \qquad Z_t = \sum_{i=1}^{t-1} e^{\,\beta\,k_t^{\top}k_i}\,,
      \qquad \beta = \beta_1 t^{\beta_2} + \beta_0\,,
  \]
  where $e^{\beta k_t^\top k_i}$ is a Gaussian kernel since keys have unit norm, with a bandwidth $\sqrt{1/\beta}$ that scales with position $t$ \citep{zhang2026memory}. This prediction is then scaled by a sigmoid gate computed from the input.

  In terms of Section~\ref{sec:principles}, the key and value extractors contribute to the input and output
  feature extractors $\phi$ (\textcolor{blue}{\ding{202}}) and $\psi$ (\textcolor{blue}{\ding{203}}), and the
  kernel regression is the inner learner $f$ (\textcolor{blue}{\ding{205}}).

        \item \emph{Transformer}: replaces the learning module (dark box) with softmax attention (\textsc{rope}). Compared with Memory Mosaics, its inner learner is not consistent: \textsc{rope}'s positional decay bounds its effective memory, and its kernel $e^{q_t^\top k_i/\sqrt{d_h}}$ has no bandwidth schedule that narrows as the context grows, so its bias does not vanish as $l\to\infty$ \citep{gyorfi2002distribution}. Its examples are also not predictive: keys and values are projections of the same token, so one attention layer retrieves a position's own representation, and forming a predictive association takes at least two layers \citep{olsson2022context,bietti2023birth}.
        \item \emph{DeltaNet}: replaces the learning module with a gated delta-rule recurrence \citep{yang2025gated}; its fixed-size state makes it inconsistent once a task outgrows that state.
    \end{itemize}
    \item[\textbf{b)}] \textbf{Modern hybrids}, interleaving DeltaNet/sliding window attention (\textsc{swa}, 2k window length)\footnote{A 4k window gives the same conclusions; see
  Appendix~\ref{apx:swa_length}.} and full attention (\textsc{nope}, no position encoding) in successive blocks. This gives two architectures: \emph{Hybrid DeltaNet} and \emph{Hybrid \textsc{swa}}. Both are configurations that frontier language models are converging on \citep{agarwal2025gpt,qiu2026design}.
    \item[\textbf{c)}] \textbf{+ architecture}.
  \begin{itemize}[leftmargin=0.4cm]
  \setlength\itemsep{-0.4em}
    \item \emph{Memory Mosaics+}: uses a separate \emph{short memory} --- a small sliding window attention module
  of size $w$ (or a mosaics or deltanet module) --- to summarize the near past below the
  \emph{long memory}.
  This allows the long memory to drop the leaky average from its keys and simply use normalized
  projections $k_t = W_k x_t / \|W_k x_t\|$.
  It also allows the long memory to skip the $s$ most recent $(k, v)$ pairs, which are strongly dependent
  on the current position in sequential data \citep{yu1994rates}, since the recent information is already captured by the short memory:
  \[\textstyle
      \frac{1}{Z_t}\sum_{i=1}^{t-s} e^{\,\beta\,k_t^{\top}k_i}\,v_i\,,
      \qquad Z_t = \sum_{i=1}^{t-s} e^{\,\beta\,k_t^{\top}k_i}\,.
  \]
  Following \citet{zhang2026memory}, $s$ is drawn uniformly from $[64, 256]$ during training and fixed at
  $64$ at inference, and $w = 256$; randomizing $s$ helps prevent the long memory from taking over the
  role of the short memory.
  
    \item \emph{Transformer+}: as Memory Mosaics+, with the same short memory, window $w$, and skip
  schedule for $s$, but with attention (\textsc{nope}, no position encoding)\footnote{Using attention with position encoding, e.g. \textsc{rope}, degrades performance. See Appendix~\ref{apx:two-stage-transformer+_ablation}.} as the long
  memory.
   \end{itemize}
    \item[\textbf{d)}] \textbf{Two-stage architecture}.
  \begin{itemize}[leftmargin=0.4cm]
  \setlength\itemsep{-0.4em}
    \item \emph{Two-stage Memory Mosaics+}: as Memory Mosaics+, but the long memory is split in two, one
  over the target task (the current document, starting from position $c$) and one over the related tasks (the earlier documents in
  the context, ending at $c-1$):\footnote{Splitting the memory does not duplicate inner-learner computation: the two sums run over disjoint parts of the context.}
 \[\textstyle
      \frac{1}{Z_t^{\mathrm{related}}}\sum_{i=1}^{c-1} e^{\,\beta\,k_t^{\top}k_i}\,v_i\;   +  \; \frac{1}{Z_t^{\mathrm{target}}}\sum_{i=c}^{t-s} e^{\,\beta\,k_t^{\top}k_i}\,v_i\,,
  \]
  where each $Z$ normalizes its own sum.
  Each sum is a weighted average of unit-norm values, so its magnitude is close to $1$ when the retrieved values agree and shrinks when they
  disagree; the two contributions therefore enter in proportion to how concordant each body of evidence
  is, an implicit form of the mixing weight $\alpha$ of Section~\ref{sec:lambda}. 
  Every structural choice in this model is one that Section~\ref{sec:inference-time-learning} calls for: a consistent
  inner learner, ($k,v$) example construction separated from prediction, and target task evidence held apart from related task evidence.

    \item \emph{Two-stage Transformer+}: as two-stage Memory Mosaics+, but with attention (\textsc{nope}) as each of the two parts of the long memory.
  \end{itemize}
  
\end{itemize}

Groups (a) and (b) are what the field currently builds. Group (c) adds a separate short memory for strongly dependent neighboring tokens, and group (d) further separates the target task from related tasks (Section~\ref{sec:lambda}).

\subsection{Experiments on rich representation and predictive disentanglement}
\label{sec:exp:rich}

 \begin{wrapfigure}{r}{0.41\linewidth}
 \vspace{-0.8em}
  \centering
  \small
  \setlength{\tabcolsep}{4pt}
  \resizebox{0.85\linewidth}{!}{
  \begin{tabular}{@{}lcccc@{}}
  \toprule
  & \multicolumn{4}{c}{Evaluation length $l^\prime$} \\
  \cmidrule(lr){2-5}
  Training length $l$ & 4k & 8k & 16k & 32k \\
  \midrule
  4k  & \cellcolor{diagred}48.55 & --- & --- & --- \\
  8k  & 48.93 & \cellcolor{diagred}48.82 & --- & --- \\
  16k & 48.97 & 49.19 & \cellcolor{diagred}49.32 & --- \\
  32k & 49.15 & 49.41 & 49.64 & \cellcolor{diagred}49.68 \\
  64k & \cellcolor{longblue}49.19 & \cellcolor{longblue}49.48
    & \cellcolor{longblue}49.84 & \cellcolor{longblue}50.00 \\
  \midrule
  $\Delta>0$ & \textcolor{blue}{\ding{52}} & \textcolor{blue}{\ding{52}} & \textcolor{blue}{\ding{52}} &
  \textcolor{blue}{\ding{52}} \\
  \bottomrule
  \end{tabular}
  }
  \captionof{table}{Memory Mosaics top-1 accuracy (\%) on SimpleWiki. Rows are the training length $l$, columns the evaluation
  length $l'$. Cells with $l<l'$ are omitted. $\Delta$ is the \textcolor{blue}{64k row}
  minus the \textcolor{red}{matched-length entry}. Every column increases downward.}
  \label{tab:rich}

  \vspace{0.6em}
  \includegraphics[width=0.8\linewidth]{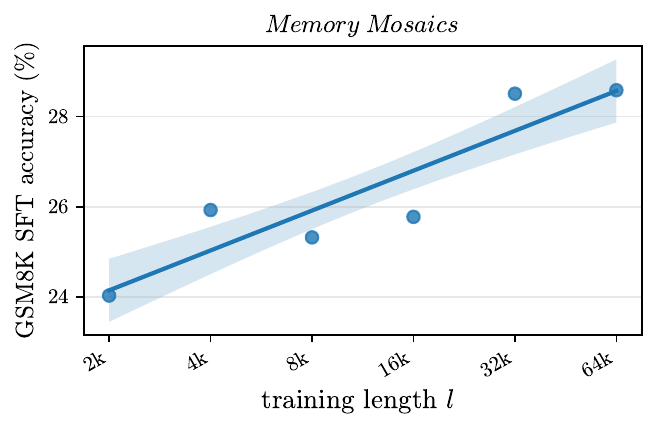}
  \captionof{figure}{\textsc{gsm8k} accuracy (\%) of Memory Mosaics after fine-tuning on math, against training length $l$. Accuracy tends
  to increase with $l$.}
  \label{fig:rich_sft}
  \vspace{-3em}
  \end{wrapfigure}

\paragraph{The positive impact of large training length $l$.}
At large $l$, $\ell_f$ vanishes for a consistent inner learner, so the training objective of
Eq.~\eqref{eq:decompose} is dominated by what the representation has discarded. Minimizing it produces a
rich representation, which provides the features that predictive disentanglement then groups. That is,
a large maximum training length $l$, holding the total amount of training data fixed, benefits
tasks other than the ones trained on.

Table~\ref{tab:rich} trains Memory Mosaics models at various maximum training lengths $l$ and
compares their performance at fixed evaluation lengths $l^\prime$ on SimpleWiki. Every column
increases downward: the model trained at $l=64$k is the best at every evaluation length,
including lengths at which the extra context it saw during training is never available.

Figure~\ref{fig:rich_sft} fine-tunes\footnote{We fine-tune each model on 262k math examples, sweep the learning rate over $\{3, 6, 9\}\times10^{-5}$ and the number of
epochs over $\{4, 6, 8\}$, and report the best configuration for each model.} the Memory Mosaics models trained at $l=2$k--$64$k (those of Table~\ref{tab:rich}, plus one at $2$k) on mathematics \citep{yu2023metamath} and evaluates them on \textsc{gsm8k}, whose problems are shorter than $1$k tokens. The result shows a positive correlation between $l$ and downstream math reasoning performance.

\paragraph{The importance of a consistent inner learner.}
The gain above depends on consistency. If the inner learner is not consistent, $\ell_f$ does not
vanish at large $l$, so $\ell_\theta$ loses the direct optimization pressure, and disentanglement loses the rich features it needs.

\begin{figure}
    % \vspace{-0.4em}
    \centering
    \includegraphics[width=0.235\linewidth]{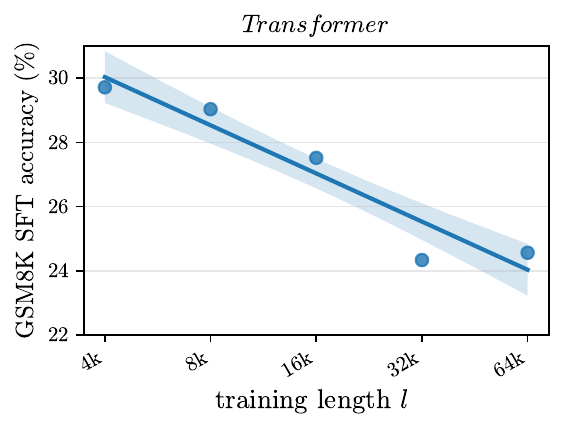}
    \includegraphics[width=0.235\linewidth]{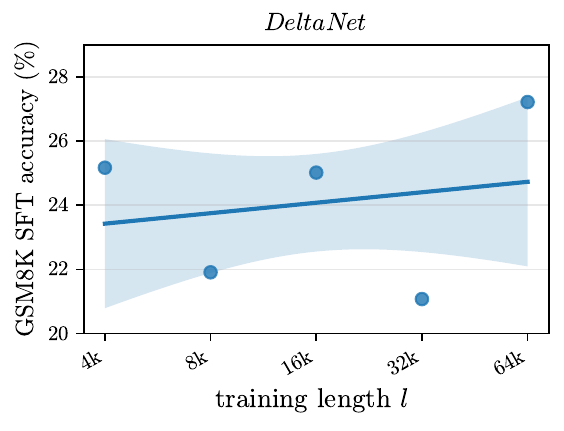}
    \includegraphics[width=0.235\linewidth]{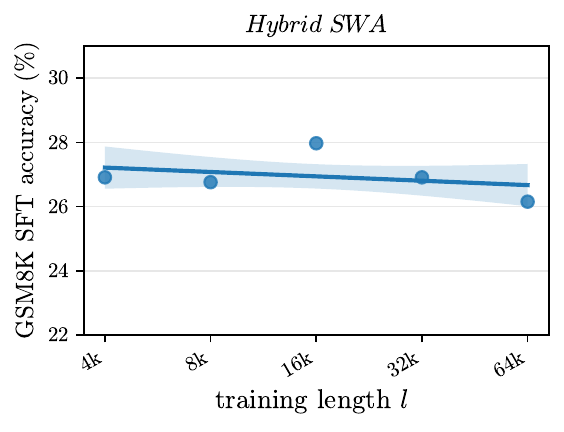}
    \includegraphics[width=0.235\linewidth]{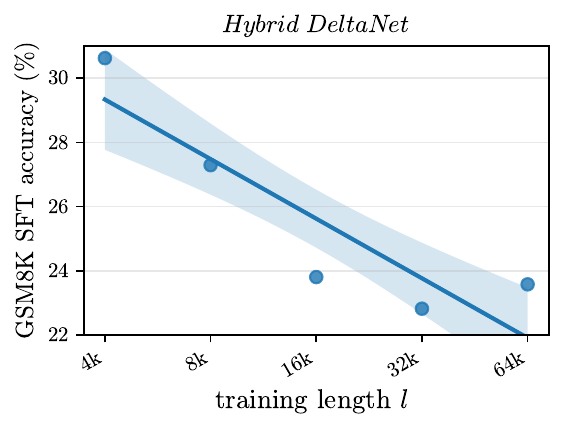}
     \caption{GSM8K accuracy (\%) after fine-tuning on math, against training length $l$, for the four architectures of Table~\ref{tab:rich-others}. Accuracy shows no systematic gain as $l$ grows, unlike Memory Mosaics (Figure~\ref{fig:rich_sft}). We do not compare accuracy \emph{quantitatively} across architectures, since it is easily altered by small engineering details. For example, Memory Mosaics has 25\% fewer learning-module
  parameters\footref{fn:params}; filling this gap by adjusting its heads raises its accuracy to 28.73\% at $l=4$k and 30.78\% at $l=64$k, while its trend with $l$ is unchanged (Appendix~\ref{apx:param_matched}).}
    \label{fig:sft_others}
    \vspace{-0.8em}
\end{figure}

Table~\ref{tab:rich-others} repeats the protocol of Table~\ref{tab:rich} for four architectures whose inner learners lack consistency: the Transformer (attention with \textsc{rope}), DeltaNet (a delta-rule recurrence), Hybrid DeltaNet, and Hybrid \textsc{swa}. The
sign flips from beneficial to harmful. None shows a systematic gain, while  Memory Mosaics in Table~\ref{tab:rich} benefits from a larger maximum training length. For all four, the best configuration is to train at or just beyond the evaluation length. The downstream results agree: fine-tuned on math in the same way as Memory Mosaics (Figure~\ref{fig:rich_sft}), none shows a systematic gain in \textsc{gsm8k} accuracy as $l$ grows (Figure~\ref{fig:sft_others}).
  \begin{table}
  \centering
  \resizebox{0.98\textwidth}{!}{%
  \small
  \setlength{\tabcolsep}{4pt}
  \begin{tabular}{@{}l  cccc  cccc  cccc cccc@{}}
  \toprule
  ~~~~~~~~~$l^\prime$
  & \multicolumn{4}{c}{\emph{Transformer}}
  & \multicolumn{4}{c}{\emph{DeltaNet}}
  & \multicolumn{4}{c}{\emph{Hybrid \textsc{swa}}}
  & \multicolumn{4}{c}{\emph{Hybrid DeltaNet}} \\
  \cmidrule(lr){2-5}\cmidrule(lr){6-9}\cmidrule(lr){10-13}\cmidrule(lr){14-17}
  $l$
  & 4k & 8k & 16k & 32k
  & 4k & 8k & 16k & 32k
  & 4k & 8k & 16k & 32k
  & 4k & 8k & 16k & 32k \\
  \midrule
  4k  & \cellcolor{diagred}48.40 & --- & --- & ---
      & \cellcolor{diagred}47.40 & --- & --- & ---
      & \cellcolor{diagred}48.19 & --- & --- & ---
      & \cellcolor{diagred}48.57 & --- & --- & --- \\
  8k  & 48.48 & \cellcolor{diagred}48.70 & --- & ---
      & 47.49 & \cellcolor{diagred}47.48 & --- & ---
      & 47.88 & \cellcolor{diagred}47.80 & --- & ---
      & 48.49 & \cellcolor{diagred}48.70 & --- & --- \\
  16k & 48.05 & 48.27 & \cellcolor{diagred}48.60 & ---
      & 47.49 & 47.43 & \cellcolor{diagred}47.58 & ---
      & 47.97 & 47.98 & \cellcolor{diagred}48.01 & ---
      & 48.15 & 48.33 & \cellcolor{diagred}48.57 & --- \\
  32k & 47.72 & 48.07 & 48.44 & \cellcolor{diagred}48.49
      & 47.27 & 47.23 & 47.31 & \cellcolor{diagred}47.15
      & 47.65 & 47.60 & 47.65 & \cellcolor{diagred}47.53
      & 46.49 & 46.51 & 46.52 & \cellcolor{diagred}46.37 \\
  64k & \cellcolor{longblue}47.51 & \cellcolor{longblue}47.85 & \cellcolor{longblue}48.23 &
  \cellcolor{longblue}48.39
      & \cellcolor{longblue}47.35 & \cellcolor{longblue}47.30 & \cellcolor{longblue}47.40 &
  \cellcolor{longblue}47.21
      & \cellcolor{longblue}47.54 & \cellcolor{longblue}47.55 & \cellcolor{longblue}47.60 &
  \cellcolor{longblue}47.46
      & \cellcolor{longblue}46.24 & \cellcolor{longblue}46.26 & \cellcolor{longblue}46.28 &
  \cellcolor{longblue}46.13 \\
  \midrule
  $\Delta>0$
      & \textcolor{red}{\ding{56}} & \textcolor{red}{\ding{56}} & \textcolor{red}{\ding{56}} &
  \textcolor{red}{\ding{56}}
      & \textcolor{red}{\ding{56}} & \textcolor{red}{\ding{56}} & \textcolor{red}{\ding{56}} &
  \textcolor{blue}{\ding{52}}
      & \textcolor{red}{\ding{56}} & \textcolor{red}{\ding{56}} & \textcolor{red}{\ding{56}} &
  \textcolor{red}{\ding{56}}
      & \textcolor{red}{\ding{56}} & \textcolor{red}{\ding{56}} & \textcolor{red}{\ding{56}} &
  \textcolor{red}{\ding{56}} \\
  \bottomrule
  \end{tabular}%
  }
  \caption{Top-1 accuracy (\%) on SimpleWiki for architectures whose inner learners lack consistency, under the same protocol as Table~\ref{tab:rich}. Red: $l=l'$ (trained at the
  evaluation length). Blue: $l=64$k. A longer max training length $l$ does not help these models
  learn new tasks; in almost every case it hurts them.}
  \label{tab:rich-others}
  \vspace{-0.4em}
  \end{table}

\paragraph{Predictive disentanglement.}
Predictive disentanglement reduces the sample complexity of a new task: the $G$-group partition of the latent is learned from the training tasks and reused, and only the prediction within each group is recomputed for the new task (Eq.~\eqref{eq:partition}). 
A consistent inner learner, which keeps the optimization pressure on the representation (Section~\ref{sec:decompose}), lets training find such a partition and therefore benefits from a large $G$; an inconsistent one cannot.

Table~\ref{tab:heads} sweeps the number of heads for Memory Mosaics, Transformer, and Hybrid \textsc{swa}.\footnote{DeltaNet and Hybrid DeltaNet are omitted, because changing the number of heads also affects the total recurrent state size. } The sign splits along consistency: Memory Mosaics gains as the number of heads rises,\footnote{The per-head gate is not the cause. See Appendix~\ref{apx:mm_mem_gate} for ablation with gate removed.} while the Transformer\footnote{The \textsc{rope} position encoding is not the cause. See Appendix~\ref{apx:mm_mem_gate} for ablation with \textsc{rope} removed.} and Hybrid \textsc{swa} lose.

\begin{table}[H]
% \vspace{-1em}
  \centering
  \resizebox{0.96\linewidth}{!}{
  \begin{tabular}{@{}l ccccc c c@{\hspace{3em}} ccccc c@{}}
  \toprule
   & \multicolumn{6}{c}{SimpleWiki top-1 (\%), $l'=4$k} & & \multicolumn{6}{c}{GSM8K accuracy (\%),
  fine-tuned} \\
  \cmidrule(lr){2-7} \cmidrule(lr){9-14}
   & \multicolumn{5}{c}{Number of groups $G$ (heads)} & & & \multicolumn{5}{c}{Number of groups $G$
  (heads)} & \\
  \cmidrule(lr){2-6} \cmidrule(lr){9-13}
  Model & 4 & 8 & 16 & 32 & 64 & $\Delta>0$ & & 4 & 8 & 16 & 32 & 64 & $\Delta>0$ \\
  \midrule
  Memory Mosaics  & \cellcolor{diagred}48.43 & 48.63 & 48.97 & 49.07 & \cellcolor{longblue}49.02 &
  \textcolor{blue}{\ding{52}} &
                  & \cellcolor{diagred}21.53 & 25.32 & 25.78 & 28.43 & \cellcolor{longblue}27.60 &
  \textcolor{blue}{\ding{52}} \\
  Transformer     & \cellcolor{diagred}48.31 & 48.38 & 48.05 & 47.92 & \cellcolor{longblue}47.08 &
  \textcolor{red}{\ding{56}} &
                  & \cellcolor{diagred}28.81 & 26.76 & 27.52 & 26.84 & \cellcolor{longblue}24.18 &
  \textcolor{red}{\ding{56}} \\
  Hybrid \textsc{swa} & \cellcolor{diagred}48.00 & 47.95 & 47.97 & 47.67 & \cellcolor{longblue}46.85 &
  \textcolor{red}{\ding{56}} &
                  & \cellcolor{diagred}26.16 & 27.67 & 27.98 & 25.55 & \cellcolor{longblue}22.82 &
  \textcolor{red}{\ding{56}} \\
  \bottomrule
  \end{tabular}}
  \caption{Effect of the number of groups $G$ (heads) at fixed model width: raising the number of heads raises $G$ and shrinks each group. All models are pre-trained
  at length $l=16$k.
  \textbf{Left:} top-1 accuracy on SimpleWiki, evaluated at length $l'=4$k.
  \textbf{Right:} \textsc{gsm8k} accuracy after fine-tuning as in Figure~\ref{fig:rich_sft}.
  $\Delta$ is (\textcolor{blue}{64 heads} $-$ \textcolor{red}{4 heads}).}
  \label{tab:heads}
  % \vspace{-1em}
  \end{table}

\subsection{Inference-time learning}
\label{sec:exp:inference}

\paragraph{Scaling the per-task length at inference.}
Consider first the simple case in which there are no related tasks, in training or at
evaluation --- only isolated ones.
The inner learner of Section~\ref{sec:principles} reads a support set of examples drawn independently from one task. A token sequence does not come this way: neighboring tokens are strongly dependent, whereas sufficiently separated blocks behave as approximately independent \citep{yu1994rates,mohri2009rademacher}. This is why in ``$+$'' architectures (Figure~\ref{fig:architecture}c) we handle neighboring tokens separately and delay the long memory to ignore neighboring tokens. 

 \begin{wrapfigure}{r}{0.5\linewidth}
    \vspace{-1.2em}
    \centering
    \includegraphics[width=0.48\linewidth]{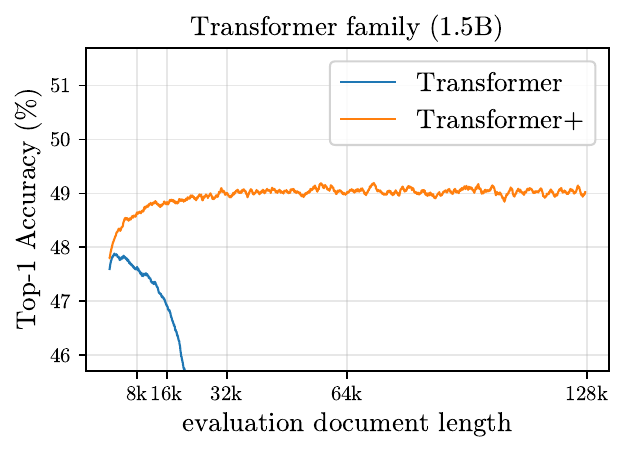}\hfill
    \includegraphics[width=0.48\linewidth]{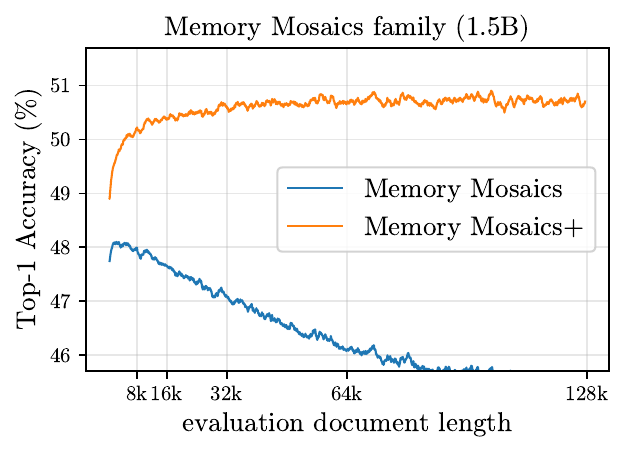}
    \caption{\textbf{Left:} Transformer vs Transformer$+$. \textbf{Right:} Memory Mosaics vs Memory Mosaics$+$. $x$-axis:
  evaluation per-document length.
    The gap widens with length.}
    \label{fig:scale_l}
    \vspace{-1em}
  \end{wrapfigure}
  
To test this, we train four models (Transformer, Transformer$+$, Memory Mosaics, Memory
Mosaics$+$) with a maximum sequence length of $16$k under a document-wise causal mask, which confines attention computation within each document even if it is very short (never across documents). We evaluate them on synthetic long documents formed by concatenating Simple English Wikipedia articles within each cluster, delimited by \texttt{\textbackslash n\textbackslash n}.

Figure~\ref{fig:scale_l} shows the result. Transformer$+$ improves on Transformer, and Memory Mosaics$+$ improves substantially on
Memory Mosaics. Both ``$+$'' models extrapolate beyond the training length. The trend does not depend on how the short memory is implemented. See Appendix~\ref{apx:ablation_short_mem} for ablation.

\paragraph{Scaling the number of related tasks at inference.}
We now return to the realistic case: few examples from the target task, and many from related
tasks. Pooling all of them is biased; Section~\ref{sec:two_predictions} prescribes modeling them separately and
    combining them dynamically, which is what the two-stage models do. The dynamic mixing term $\alpha$ is implicitly controlled via the magnitudes of the two parts. See Appendix~\ref{apx:two-stage_magnitudes} for details. 

\begin{figure}[H]
\vspace{-0.4em}
    \centering
    \includegraphics[width=0.235\linewidth]{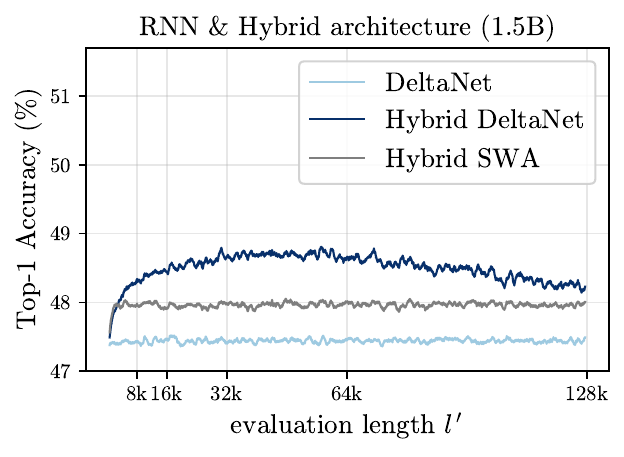}
    \includegraphics[width=0.235\linewidth]{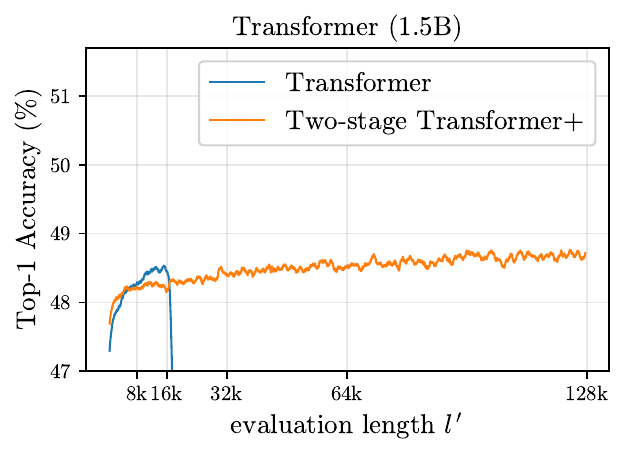}
    \includegraphics[width=0.235\linewidth]{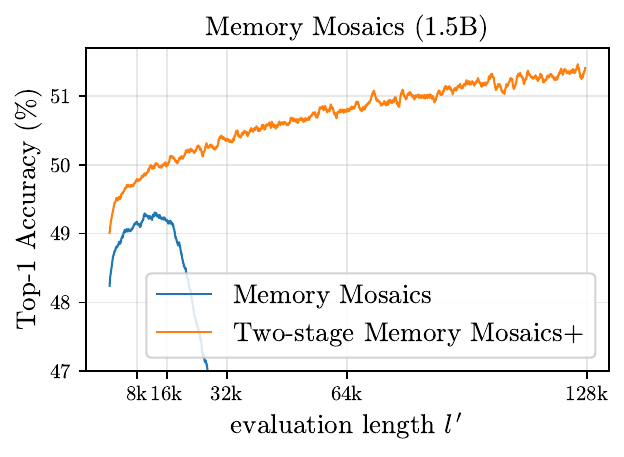}
    \includegraphics[width=0.24\linewidth]{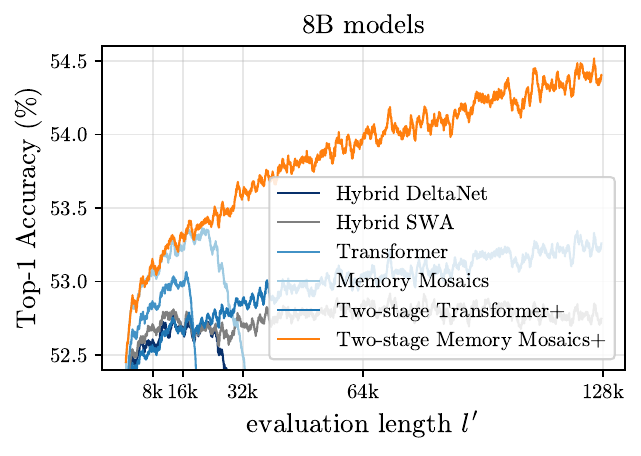}
    \caption{Accuracy as the number of related tasks in the context grows. Training length $l=16$k. At $1.5$B:
    \textbf{1)} RNN and hybrid architectures; \textbf{2)} Transformer versus two-stage
    Transformer$+$; \textbf{3)} Memory Mosaics versus two-stage Memory Mosaics$+$. Panel
    \textbf{4)} repeats the comparison at $8$\textsc{b}.}
    \label{fig:two-stage_vs_one-stage}
    \vspace{-1.3em}
\end{figure}

Figure~\ref{fig:two-stage_vs_one-stage} shows the result. One-stage models peak around their
training length and then decline or flatten. Two-stage Transformer$+$ keeps improving, though
slowly; two-stage Memory Mosaics$+$ improves quickly. The rightmost panel repeats the comparison
at $\sim$$8$\textsc{b} scale, over established architectures, hybrids, and two-stage $+$ variants, and every curve keeps
its shape --- even though each member of the Memory Mosaics family is smaller than its counterpart\footref{fn:params}.
Capacity is not the missing ingredient; structure is.

\section{Conclusion}

Open-world AI asks for a machine that takes up tasks it was not built for --- the first requirement, \emph{inference-time learning}. 
Modeling the problem with tasks that share latent factors, and decomposing the training objective, this work derives two more requirements, a \emph{rich representation} and a \emph{predictive disentanglement}, and two predictions: that the inner learner must be consistent, and that the target task and the many related tasks in a long context must be handled separately. The experiments test each qualitatively, asking whether a model has the property or lacks it. 
Models with these properties improve as the training length $l$ or the number of related tasks grows, while the others degrade.

\bibliographystyle{iclr2027_conference}
\bibliography{refs}

@inproceedings{finn2017model,
	title        = {Model-agnostic meta-learning for fast adaptation of deep networks},
	author       = {Finn, Chelsea and Abbeel, Pieter and Levine, Sergey},
	year         = 2017,
	booktitle    = {International conference on machine learning},
	pages        = {1126--1135},
	organization = {PMLR}
}

@inproceedings{koh2021wilds,
	title        = {Wilds: A benchmark of in-the-wild distribution shifts},
	author       = {Koh, Pang Wei and Sagawa, Shiori and Marklund, Henrik and Xie, Sang Michael and Zhang, Marvin and Balsubramani, Akshay and Hu, Weihua and Yasunaga, Michihiro and Phillips, Richard Lanas and Gao, Irena and others},
	year         = 2021,
	booktitle    = {International Conference on Machine Learning},
	pages        = {5637--5664},
	organization = {PMLR}
}

@inproceedings{zhang2022rich,
	title        = {Rich feature construction for the optimization-generalization dilemma},
	author       = {Zhang, Jianyu and Lopez-Paz, David and Bottou, L{\'e}on},
	year         = 2022,
	booktitle    = {International Conference on Machine Learning},
	pages        = {26397--26411},
	organization = {PMLR}
}

@inproceedings{chen2020simple,
	title        = {A simple framework for contrastive learning of visual representations},
	author       = {Chen, Ting and Kornblith, Simon and Norouzi, Mohammad and Hinton, Geoffrey},
	year         = 2020,
	booktitle    = {International conference on machine learning},
	pages        = {1597--1607},
	organization = {PMLR}
}

@inproceedings{zhang2023learning,
	title        = {Learning useful representations for shifting tasks and distributions},
	author       = {Zhang, Jianyu and Bottou, L{\'e}on},
	year         = 2023,
	booktitle    = {International Conference on Machine Learning},
	pages        = {40830--40850},
	organization = {PMLR}
}

@article{vapnik1991principles,
	title        = {Principles of risk minimization for learning theory},
	author       = {Vapnik, Vladimir},
	year         = 1991,
	journal      = {Advances in neural information processing systems},
	volume       = 4
}

@article{scholkopf2021toward,
	title        = {Toward causal representation learning},
	author       = {Sch{\"o}lkopf, Bernhard and Locatello, Francesco and Bauer, Stefan and Ke, Nan Rosemary and Kalchbrenner, Nal and Goyal, Anirudh and Bengio, Yoshua},
	year         = 2021,
	journal      = {Proceedings of the IEEE},
	publisher    = {IEEE},
	volume       = 109,
	number       = 5,
	pages        = {612--634}
}

@article{chen2023towards,
	title        = {Towards Understanding Feature Learning in Out-of-Distribution Generalization},
	author       = {Chen, Yongqiang and Huang, Wei and Zhou, Kaiwen and Bian, Yatao and Han, Bo and Cheng, James},
	year         = 2023,
	journal      = {arXiv preprint arXiv:2304.11327}
}

@inproceedings{recht2019imagenet,
	title        = {Do imagenet classifiers generalize to imagenet?},
	author       = {Recht, Benjamin and Roelofs, Rebecca and Schmidt, Ludwig and Shankar, Vaishaal},
	year         = 2019,
	booktitle    = {International conference on machine learning},
	pages        = {5389--5400},
	organization = {PMLR}
}

@inproceedings{bardes2022vicreg,
	title        = {{VICR}eg: Variance-Invariance-Covariance Regularization for Self-Supervised Learning},
	author       = {Adrien Bardes and Jean Ponce and Yann LeCun},
	year         = 2022,
	booktitle    = {International Conference on Learning Representations}
}

@article{li2024long,
	title        = {Long-context llms struggle with long in-context learning},
	author       = {Li, Tianle and Zhang, Ge and Do, Quy Duc and Yue, Xiang and Chen, Wenhu},
	year         = 2024,
	journal      = {arXiv preprint arXiv:2404.02060}
}

@unpublished{mccarthy1955proposal,
  author    = {McCarthy, John and Minsky, Marvin and Rochester, Nathaniel and Shannon, Claude E.},
  title     = {A Proposal for the Dartmouth Summer Research Project on Artificial Intelligence},
  note      = {Unpublished manuscript, Dartmouth College},
  month     = {August},
  year      = {1955}
}

@article{samuel1959some,
	title        = {Some studies in machine learning using the game of checkers},
	author       = {Samuel, Arthur L},
	year         = 1959,
	journal      = {IBM Journal of research and development},
	publisher    = {IBM},
	volume       = 3,
	number       = 3,
	pages        = {210--229}
}

@article{tesauro1995temporal,
	title        = {Temporal difference learning and TD-Gammon},
	author       = {Tesauro, Gerald},
	year         = 1995,
	journal      = {Communications of the ACM},
	volume       = 38,
	number       = 3,
	pages        = {58--68}
}

@article{silver2016mastering,
	title        = {Mastering the game of Go with deep neural networks and tree search},
	author       = {Silver, David and Huang, Aja and Maddison, Chris J and Guez, Arthur and Sifre, Laurent and Van Den Driessche, George and Schrittwieser, Julian and Antonoglou, Ioannis and Panneershelvam, Veda and Lanctot, Marc and others},
	year         = 2016,
	journal      = {nature},
	publisher    = {Nature Publishing Group},
	volume       = 529,
	number       = 7587,
	pages        = {484--489}
}

@article{matan1991multi,
	title        = {Multi-digit recognition using a space displacement neural network},
	author       = {Matan, Ofer and Burges, Christopher J and LeCun, Yann and Denker, John},
	year         = 1991,
	journal      = {Advances in neural information processing systems},
	volume       = 4
}

@inproceedings{taigman2014deepface,
	title        = {Deepface: Closing the gap to human-level performance in face verification},
	author       = {Taigman, Yaniv and Yang, Ming and Ranzato, Marc'Aurelio and Wolf, Lior},
	year         = 2014,
	booktitle    = {Proceedings of the IEEE conference on computer vision and pattern recognition},
	pages        = {1701--1708}
}

@inproceedings{zhang-2025,
title={Memory Mosaics},
author={Jianyu Zhang and Niklas Nolte and Ranajoy Sadhukhan and Beidi Chen and L\'eon Bottou},
booktitle={The Thirteenth International Conference on Learning Representations},
year={2025},
url={https://openreview.net/forum?id=IiagjrJNwF}
}

@article{olsson2022context,
  title={In-context learning and induction heads},
  author={Olsson, Catherine and Elhage, Nelson and Nanda, Neel and Joseph, Nicholas and DasSarma, Nova and Henighan, Tom and Mann, Ben and Askell, Amanda and Bai, Yuntao and Chen, Anna and others},
  journal={arXiv preprint arXiv:2209.11895},
  year={2022}
}

@article{bietti2023birth,
  title={Birth of a transformer: A memory viewpoint},
  author={Bietti, Alberto and Cabannes, Vivien and Bouchacourt, Diane and Jegou, Herve and Bottou, Leon},
  journal={Advances in Neural Information Processing Systems},
  volume={36},
  pages={1560--1588},
  year={2023}
}

@article{lecun2022path,
  title={A path towards autonomous machine intelligence version 0.9. 2, 2022-06-27},
  author={LeCun, Yann},
  journal={Open Review},
  volume={62},
  number={1},
  pages={1--62},
  year={2022}
}

@inproceedings{liu2025decade,
  title={A decade's battle on dataset bias: Are we there yet?},
  author={Liu, Zhuang and He, Kaiming},
  booktitle={International Conference on Learning Representations},
  volume={2025},
  pages={41433--41449},
  year={2025}
}

@INPROCEEDINGS{Torralba11,
  author={Torralba, Antonio and Efros, Alexei A.},
  booktitle={CVPR 2011}, 
  title={Unbiased look at dataset bias}, 
  year={2011},
  volume={},
  number={},
  pages={1521-1528},
  doi={10.1109/CVPR.2011.5995347}}

@article{Maurer2016,
  author  = {Andreas Maurer and Massimiliano Pontil and Bernardino Romera-Paredes},
  title   = {The Benefit of Multitask Representation Learning},
  journal = {Journal of Machine Learning Research},
  year    = {2016},
  volume  = {17},
  number  = {81},
  pages   = {1--32},
  url     = {http://jmlr.org/papers/v17/15-242.html}
}

@book{gelman2007data,
  title={Data analysis using regression and multilevel/hierarchical models},
  author={Gelman, Andrew and Hill, Jennifer},
  volume={1},
  year={2007},
  publisher={Cambridge university press}
}

@article{zech2018variable,
  title={Variable generalization performance of a deep learning model to detect pneumonia in chest radiographs: a cross-sectional study},
  author={Zech, John R and Badgeley, Marcus A and Liu, Manway and Costa, Anthony B and Titano, Joseph J and Oermann, Eric Karl},
  journal={PLoS medicine},
  volume={15},
  number={11},
  pages={e1002683},
  year={2018},
  publisher={Public Library of Science San Francisco, CA USA}
}

@inproceedings{zadrozny2004learning,
  title={Learning and evaluating classifiers under sample selection bias},
  author={Zadrozny, Bianca},
  booktitle={Proceedings of the twenty-first international conference on Machine learning},
  pages={114},
  year={2004}
}

@article{heckman1979sample,
  title={Sample selection bias as a specification error},
  author={Heckman, James J},
  journal={Econometrica: Journal of the econometric society},
  pages={153--161},
  year={1979},
  publisher={JSTOR}
}

@book{gyorfi2002distribution,
  title={A distribution-free theory of nonparametric regression},
  author={Gy{\"o}rfi, L{\'a}szl{\'o} and Kohler, Michael and Krzy{\.z}ak, Adam and Walk, Harro},
  year={2002},
  publisher={Springer}
}

@article{yang2025qwen3,
  title={Qwen3 technical report},
  author={Yang, An and Li, Anfeng and Yang, Baosong and Zhang, Beichen and Hui, Binyuan and Zheng, Bo and Yu, Bowen and Gao, Chang and Huang, Chengen and Lv, Chenxu and others},
  journal={arXiv preprint arXiv:2505.09388},
  year={2025}
}

@misc{benallal2024cosmopedia,
  author = {Ben Allal, Loubna and Lozhkov, Anton and Penedo, Guilherme and Wolf, Thomas and von Werra, Leandro},
  title = {Cosmopedia},
  month = {February},
  year = 2024,
  url = {https://huggingface.co/datasets/HuggingFaceTB/cosmopedia}
}

@inproceedings{su2025nemotron,
  title={Nemotron-cc: Transforming common crawl into a refined long-horizon pretraining dataset},
  author={Su, Dan and Kong, Kezhi and Lin, Ying and Jennings, Joseph and Norick, Brandon and Kliegl, Markus and Patwary, Mostofa and Shoeybi, Mohammad and Catanzaro, Bryan},
  booktitle={Proceedings of the 63rd Annual Meeting of the Association for Computational Linguistics (Volume 1: Long Papers)},
  pages={2459--2475},
  year={2025}
}

@misc{sharegpt_vicuna_unfiltered,
  author       = {Vicuna},
  title        = {ShareGPT Vicuna Unfiltered},
  year         = {2023},
  publisher    = {Hugging Face},
  howpublished = {\url{https://huggingface.co/datasets/anon8231489123/ShareGPT_Vicuna_unfiltered}}
}

@inproceedings{azerbayev2024llemma,
  title={Llemma: An open language model for mathematics},
  author={Azerbayev, Zhangir and Schoelkopf, Hailey and Paster, Keiran and Dos Santos, Marco and McAleer, Stephen and Jiang, Qiaochu and Deng, Jia and Biderman, Stella R and Welleck, Sean},
  booktitle={International Conference on Learning Representations},
  volume={2024},
  pages={40622--40649},
  year={2024}
}

@article{nussbaum2024nomic,
  title={Nomic embed: Training a reproducible long context text embedder},
  author={Nussbaum, Zach and Morris, John X and Duderstadt, Brandon and Mulyar, Andriy},
  journal={arXiv preprint arXiv:2402.01613},
  year={2024}
}

@article{tripuraneni2020theory,
  title={On the theory of transfer learning: The importance of task diversity},
  author={Tripuraneni, Nilesh and Jordan, Michael and Jin, Chi},
  journal={Advances in neural information processing systems},
  volume={33},
  pages={7852--7862},
  year={2020}
}

@article{du2020few,
  title={Few-shot learning via learning the representation, provably},
  author={Du, Simon S and Hu, Wei and Kakade, Sham M and Lee, Jason D and Lei, Qi},
  journal={arXiv preprint arXiv:2002.09434},
  year={2020}
}

@techreport{deepseek2026v41flash,
  author      = {DeepSeek-AI},
  title       = {DeepSeek-V4.1-Flash: Pushing the Limits of KV Cache Compression},
  institution = {DeepSeek-AI},
  year        = {2026},
  url         = {https://huggingface.co/deepseek-ai/DeepSeek-V4.1-Flash}
}

@article{stone1985additive,
  title={Additive regression and other nonparametric models},
  author={Stone, Charles J},
  journal={The annals of Statistics},
  pages={689--705},
  year={1985},
  publisher={JSTOR}
}

@article{oord2018representation,
  title={Representation learning with contrastive predictive coding},
  author={Oord, Aaron van den and Li, Yazhe and Vinyals, Oriol},
  journal={arXiv preprint arXiv:1807.03748},
  year={2018}
}

@misc{manu2023gutenberg,
title        = {Project Gutenberg},
author       = {{Manu}},
year         = {2023},
publisher    = {Hugging Face},
howpublished = {\url{https://huggingface.co/datasets/manu/project_gutenberg}},
}

@inproceedings{mohri2009rademacher,
 author = {Mohri, Mehryar and Rostamizadeh, Afshin},
 booktitle = {Advances in Neural Information Processing Systems},
 editor = {D. Koller and D. Schuurmans and Y. Bengio and L. Bottou},
 pages = {},
 publisher = {Curran Associates, Inc.},
 title = {Rademacher Complexity Bounds for Non-I.I.D. Processes},
 volume = {21},
 year = {2008}
}

@article{yu1994rates,
  title={Rates of convergence for empirical processes of stationary mixing sequences},
  author={Yu, Bin},
  journal={The Annals of Probability},
  pages={94--116},
  year={1994},
  publisher={JSTOR}
}

@article{zhang2026memory,
  title={Memory Mosaics at scale},
  author={Zhang, Jianyu and Bottou, L{\'e}on},
  journal={Advances in Neural Information Processing Systems},
  volume={38},
  pages={38929--38956},
  year={2026}
}

@article{yu2023metamath,
  title={MetaMath: Bootstrap Your Own Mathematical Questions for Large Language Models},
  author={Yu, Longhui and Jiang, Weisen and Shi, Han and Yu, Jincheng and Liu, Zhengying and Zhang, Yu and Kwok, James T and Li, Zhenguo and Weller, Adrian and Liu, Weiyang},
  journal={arXiv preprint arXiv:2309.12284},
  year={2023}
}

@article{baxter2000model,
  title={A model of inductive bias learning},
  author={Baxter, Jonathan},
  journal={Journal of artificial intelligence research},
  volume={12},
  pages={149--198},
  year={2000}
}

@inproceedings{garnelo2018conditional,
  title={Conditional neural processes},
  author={Garnelo, Marta and Rosenbaum, Dan and Maddison, Christopher and Ramalho, Tiago and Saxton, David and Shanahan, Murray and Teh, Yee Whye and Rezende, Danilo and Eslami, SM Ali},
  booktitle={International conference on machine learning},
  pages={1704--1713},
  year={2018},
  organization={PMLR}
}

@article{kim2019attentive,
  title={Attentive neural processes},
  author={Kim, Hyunjik and Mnih, Andriy and Schwarz, Jonathan and Garnelo, Marta and Eslami, Ali and Rosenbaum, Dan and Vinyals, Oriol and Teh, Yee Whye},
  journal={arXiv preprint arXiv:1901.05761},
  year={2019}
}

@article{grant2018recasting,
  title={Recasting gradient-based meta-learning as hierarchical bayes},
  author={Grant, Erin and Finn, Chelsea and Levine, Sergey and Darrell, Trevor and Griffiths, Thomas},
  journal={arXiv preprint arXiv:1801.08930},
  year={2018}
}

@inproceedings{yang2025gated,
  title={Gated delta networks: Improving mamba2 with delta rule},
  author={Yang, Songlin and Kautz, Jan and Hatamizadeh, Ali},
  booktitle={International Conference on Learning Representations},
  volume={2025},
  pages={29687--29707},
  year={2025}
}

@article{agarwal2025gpt,
  title={gpt-oss-120b \& gpt-oss-20b model card},
  author={Agarwal, Sandhini and Ahmad, Lama and Ai, Jason and Altman, Sam and Applebaum, Andy and Arbus, Edwin and Arora, Rahul K and Bai, Yu and Baker, Bowen and Bao, Haiming and others},
  journal={arXiv preprint arXiv:2508.10925},
  year={2025}
}

@article{qiu2026design,
  title={On the design of {qwen3.8-next} architecture: Evaluation, efficiency, and training stability},
  author={Qiu, Zihan and Wang, Zekun and Li, Xiao and Li, Yanpeng and Xu, Yang and Wang, Yixuan and Zhang, Huaqing and Men, Rui and Mao, Bochao and Zhang, Chengruidong and others},
  journal={arXiv preprint arXiv:2608.30320},
  year={2026}
}
\clearpage

\pagebreak
\appendix
\hrule
\begin{center}
\LARGE Open-world AI: a Tale of Three Requirements
\end{center}

\begin{center}
\large Supplementary Material
\end{center}

\hrule
\vskip 1cm
\section{A linear Gaussian case study}
\label{apx:linear_gaussian}

This section makes \(\mathbb{E}_{\tau}[\ell_\theta(l)-\sigma^2_\tau]\) of Eq.~\eqref{eq:decompose} explicit in a case where every quantity can be computed, and shows what a \emph{rich representation} is there.

\paragraph{Data generation.}
Each task \(\tau\) is a parameter \(\omega\in\mathbb{R}^d\):
\begin{align*}
    y = \omega^\top x + \epsilon, \qquad x \sim \mathcal{N}(0, \Sigma_x), \qquad \epsilon \sim \mathcal{N}(0, \sigma_\epsilon^2), \qquad   \omega \sim p(\omega),
\end{align*}
with \(\omega\), \(x\) and \(\epsilon\) mutually independent, \(\Sigma_\omega := \mathbb{E}[\omega\omega^\top]\) finite, and \(\mathrm{rank}(\Sigma_\omega)\ge m\), \(m\le d\). The irreducible error is the same for every task, \(\sigma^2_\tau=\sigma^2_\epsilon\).

The \textbf{input feature extractor} is a rank-\(m\) linear bottleneck, \(\phi=V\in\mathbb{R}^{m\times d}\), \(z=Vx\in\mathbb{R}^m\). 
The \textbf{output feature extractor} and \textbf{decoder} are set to identity.
The \textbf{inner learner} \(f\) is ordinary least squares (OLS) with weight \(\hat\mu\in\mathbb{R}^m\) and prediction \(\hat y^{*}=\hat\mu^\top z^{*}\). OLS is chosen because at \(l\to\infty\) it attains the Bayes prediction given \(z\): \(\ell_f(\theta,\infty)=0\) exactly, so the whole of the remaining error is \(\ell_\theta\).

Work in the whitened coordinates \(\tilde x = \Sigma_x^{-1/2}x \sim \mathcal{N}(0,I)\), \(\tilde\omega = \Sigma_x^{1/2}\omega\) and \(\tilde V = V\Sigma_x^{1/2}\in\mathbb{R}^{m\times d}\), so that \(y = \tilde\omega^\top\tilde x + \epsilon\) and \(z = \tilde V\tilde x\). Write
\[
  M := \mathbb{E}\bigl[\tilde\omega\tilde\omega^\top\bigr]
     = \Sigma_x^{1/2}\,\Sigma_\omega\,\Sigma_x^{1/2}
\]
for the whitened task-signal covariance, and \(P_{\tilde V} = \tilde V^\top(\tilde V\tilde V^\top)^+\tilde V\) for the orthogonal projection onto \(\mathrm{row}(\tilde V)\), where \(^+\) is the Moore--Penrose pseudoinverse.

\paragraph{What the representation discards.}
\textbf{Proposition 1.} \emph{With OLS as the inner learner,}
\[
    \mathbb{E}_{\tau}\bigl[\ell_\theta(\infty) - \sigma^2_\tau\bigr]
    \;=\; \mathbb{E}_\omega\bigl[\tilde\omega^\top(I-P_{\tilde V})\tilde\omega\bigr]
    \;=\; \mathrm{tr}\bigl(M\,(I-P_{\tilde V})\bigr).
\]
This is the expected squared task signal lying in the whitened directions \textbf{orthogonal to the row space of \(\tilde V\)}. It does not depend on \(l\): no amount of task data recovers it, only a better \(V\) does. This is the quantity Section~\ref{sec:decompose} calls the information the representation has permanently discarded.

\begin{coloredproof}
\begin{proof}
\(P_{\tilde V}\) is symmetric and idempotent with \(\mathrm{rank}(P_{\tilde V})=\mathrm{rank}(\tilde V)\), and \((I-P_{\tilde V})\tilde x \perp \mathrm{row}(\tilde V)\).

As \(l\to\infty\) the OLS normal equations converge, \((Z^\top Z)/l \to \mathbb{E}[zz^\top]\) and \(Z^\top y/l \to \mathbb{E}[zy]\), so for a task \(\omega\),
\[
    \hat\mu_\omega = (Z^\top Z)^{+} Z^\top y
    \;\longrightarrow\; \mathbb{E}[zz^\top]^{+}\,\mathbb{E}[zy]
    = (V\Sigma_xV^\top)^+V\Sigma_x\omega ,
\]
giving the prediction and the residual
\begin{align*}
    \hat y^*_\infty &= \hat{\mu}_\omega^\top Vx^* = \tilde\omega^\top P_{\tilde V}\tilde x^* ,\\
    y^* - \hat y^*_\infty &= \tilde\omega^\top\tilde x^* + \epsilon  - \hat y^*_\infty
    = \tilde\omega^\top\underbrace{(I-P_{\tilde V})\tilde x^*}_{\perp\;\mathrm{row}(\tilde V)} + \epsilon .
\end{align*}
Squaring and taking expectations,
\begin{align*}
    \mathbb{E}_{\omega, x^*, \epsilon}\bigl[(y^* - \hat y^*_\infty)^2\bigr]
    &= \mathbb{E}\bigl[\bigl(\tilde\omega^\top(I-P_{\tilde V})\tilde x^*\bigr)^2\bigr]
     + 2\,\underbrace{\mathbb{E}\bigl[\tilde\omega^\top(I-P_{\tilde V})\tilde x^*\,\epsilon\bigr]}_{=\,0}
     + \underbrace{\mathbb{E}[\epsilon^2]}_{=\,\sigma^2_\epsilon} \\
    &= \mathbb{E}_\omega\bigl[\tilde\omega^\top(I-P_{\tilde V})\underbrace{\mathbb{E}[\tilde x^*\tilde x^{*\top}]}_{=\,I}(I-P_{\tilde V})\tilde\omega\bigr] + \sigma^2_\epsilon \\
    &= \mathbb{E}_\omega\bigl[\tilde\omega^\top(I-P_{\tilde V})\tilde\omega\bigr] + \sigma^2_\epsilon
     \;=\; \mathrm{tr}\bigl(M\,(I-P_{\tilde V})\bigr) + \sigma^2_\epsilon .
\end{align*}
The cross term vanishes because \(\epsilon\) is zero-mean and independent of \((\omega,x^{*})\). The second-to-last equality uses idempotency, \((I-P_{\tilde V})^2 = I-P_{\tilde V}\), and the last uses \(\mathbb{E}_\omega[\tilde\omega^\top A\tilde\omega] = \mathrm{tr}(\mathbb{E}[\tilde\omega\tilde\omega^\top]A)\) for symmetric \(A\). Subtracting \(\sigma^2_\tau = \sigma^2_\epsilon\) gives the claim.
\end{proof}
\end{coloredproof}

\paragraph{What the best representation retains.}
\textbf{Proposition 2.} \emph{Let \(M = U\Lambda U^\top\) with \(\Lambda=\mathrm{diag}(\lambda_1\ge\cdots\ge\lambda_d\ge 0)\). Then}
\[
    \min_{V\in\mathbb{R}^{m\times d}} \;\mathrm{tr}\bigl(M(I-P_{\tilde V})\bigr) \;=\; \sum_{i>m}\lambda_i ,
\]
\emph{attained by any \(V\) whose whitened row space \(\tilde V = V\Sigma_x^{1/2}\) spans the top-\(m\) eigenvectors of \(M\).}

\begin{coloredproof}
\begin{proof}
Since \(\mathrm{tr}(M(I-P_{\tilde V})) = \mathrm{tr}(M) - \mathrm{tr}(MP_{\tilde V})\), minimizing the former is maximizing \(\mathrm{tr}(MP_{\tilde V})\), subject to \(\mathrm{rank}(P_{\tilde V}) = \mathrm{rank}(\tilde V)\le m\).

Put \(Q := U^\top P_{\tilde V} U\). It is again an orthogonal projection --- symmetric and idempotent, \(Q^\top = Q = Q^2\) --- of the same rank. With \(w_i := Q_{ii}\),
\[\mathrm{tr}(MP_{\tilde V}) = \mathrm{tr}(U\Lambda U^\top P_{\tilde V}) = \mathrm{tr}(\Lambda Q) = \sum_{i=1}^d \lambda_i w_i .\]
Because \(Q\) is an orthogonal projection, \(w_i\in[0,1]\) and \(\sum_i w_i = \mathrm{tr}(Q) = \mathrm{tr}(P_{\tilde V}UU^\top) = \mathrm{rank}(P_{\tilde V}) \le m\). The sum \(\sum_i\lambda_i w_i\) is therefore at most \(\sum_{i\le m}\lambda_i\), attained at \(w_1=\cdots=w_m=1\) and \(w_{m+1}=\cdots=w_d=0\), that is \(Q = \mathrm{diag}(I_m,0)\) and \(P_{\tilde V} = U\,\mathrm{diag}(I_m,0)\,U^\top\). This is exactly the projection onto the span of the top-\(m\) eigenvectors of \(M\), and the minimum value is \(\mathrm{tr}(M)-\sum_{i\le m}\lambda_i = \sum_{i>m}\lambda_i\).
\end{proof}
\end{coloredproof}

\paragraph{Why a second principle is needed.}
Proposition 2 determines a \emph{subspace}, not a basis: replacing \(\tilde V\) by \(R\tilde V\) for any invertible \(R\in\mathbb{R}^{m\times m}\) leaves \(P_{\tilde V}\) unchanged, and with it the loss. The large-\(l\) pressure of Section~\ref{sec:decompose} therefore fixes what the representation must retain and says nothing about how those \(m\) directions are arranged into groups. Choosing the arrangement is the separate job of predictive disentanglement at small \(l\) --- the same invariance that makes \(\bar y^{*}\) independent of how \(\psi\) organizes \(\gamma\).

\section{Experiment details}
\label{apx:experiment_details}

\subsection{Architecture}

Every model shares the layout of Section~\ref{sec:exp:models}: a token embedding, a stack of blocks each pairing a learning module with a SwiGLU feed-forward network, and a linear decoder. Embeddings are untied. Table~\ref{tab:apx_arch} gives the settings held fixed at each of the two scales.

\begin{table}[H]
\centering
\small
\begin{tabular}{@{}lrr@{}}
\toprule
 & $1.5$B & $8$B \\
\midrule
model dimension $d$        & 2\,048  & 4\,096 \\
layers                     & 24      & 32 \\
head dimension             & 128     & 128 \\
heads                      & 16      & 32 \\
\textsc{ffn} hidden dimension & 5\,632 $(2.75\,d)$ & 14\,336 $(3.5\,d)$ \\
vocabulary (Mistral-7B-v0.1)  & 32\,000 & 32\,000 \\
\bottomrule
\end{tabular}
\caption{Architecture settings shared by all models at each scale. The feed-forward width follows the Llama SwiGLU convention (see text), and the input and output embeddings are not tied.}
\label{tab:apx_arch}
\end{table}

The feed-forward width follows the Llama SwiGLU convention: a nominal $4d$ expansion is multiplied by $2/3$ to compensate for SwiGLU's three weight matrices, then by a size-dependent factor ($1.0$ at $1.5$B, $1.3$ at $8$B) and rounded up to a multiple of $256$. The hidden dimension is therefore $2.75\,d$ and $3.5\,d$, not $4d$.

\paragraph{Position encoding.}

\textsc{rope} splits each head into $d_h/2$ channel pairs and rotates pair $i$ of the query and key by an angle $t\,\omega_i$ at position $t$, with $\omega_i = \theta^{-2i/d_h}$ and $i=0,\ldots,d_h/2-1$, where $d_h$ is the head dimension. 

Wherever \textsc{rope} is used, its base $\theta$ is scaled linearly with the span the layer attends over, at a fixed ratio $\theta/\text{context length}\approx122$.

This applies to the short sliding window ($256$), to the sliding-window blocks of the hybrids ($2$k), and to the Transformer of Table~\ref{tab:rich-others} at each of its training lengths. The Transformer baseline is therefore given a correctly rescaled \textsc{rope} at every $l$, so its behaviour in Table~\ref{tab:rich-others} is not an artifact of a mistuned $\theta$.

\paragraph{Hybrids.}
The hybrids interleave local and global blocks $1{:}1$: even-numbered layers are local, odd-numbered layers are full attention without position encoding, so layer $0$ is local and the last layer is full attention. That gives $12$ of each at $1.5$B and $16$ of each at $8$B. The local block differs by family: \emph{Hybrid \textsc{swa}} uses sliding-window attention with \textsc{rope}, while \emph{Hybrid DeltaNet} uses a recurrence and therefore carries no position encoding at all.

\subsection{Training}

All models are trained with AdamW ($\beta_1=0.9$, $\beta_2=0.95$, $\epsilon=10^{-8}$) and weight decay $0.1$, peak learning rate $3\times10^{-4}$ with $200$ warmup steps and a cosine decay to $1\%$ of peak, and gradient clipping at $1.0$. Weight decay is not applied to the learned bandwidth and damping coefficients, whose optima are not at zero. Training is in \texttt{bf16} under \textsc{fsdp}. Every run uses the same seed; no pre-training hyper-parameter is tuned per model.

\paragraph{Token budget.}
The $1.5$B models train for $4000$ steps and the $8$B models for $8000$, at a global batch of $2^{22}=4.19$M tokens per step. For the training-length sweeps of Tables~\ref{tab:rich} and~\ref{tab:rich-others}, the number of \textsc{gpu}s and the per-device batch are chosen together so that this global batch is identical at every $l$:

\begin{center}
\small
\begin{tabular}{@{}lccccc@{}}
\toprule
training length $l$ & $4$k & $8$k & $16$k & $32$k & $64$k \\
\midrule
\textsc{gpu}s          & 256 & 256 & 256 & 128 & 64 \\
tokens / step       & \multicolumn{5}{c}{$4.19$M} \\
\bottomrule
\end{tabular}
\end{center}

Every model in that comparison therefore sees the same number of tokens in the same number of optimizer steps, and only the length of the sequences differs.

\paragraph{Data.}
Documents are drawn from six sources in the proportions of Table~\ref{tab:apx_mixture}. The percentages are shares of \emph{documents}, not of tokens. Cosmopedia and NemotronCC-high mainly contain short documents, so their token contribution is well below their $94\%$ share of the documents.

\begin{table}[H]
\centering
\small
\begin{tabular}{@{}lr@{}}
\toprule
Source & \% of documents \\
\midrule
Cosmopedia \citep{benallal2024cosmopedia}        & $71.17$ \\
NemotronCC-high \citep{su2025nemotron}           & $22.89$ \\
open-web-math \citep{azerbayev2024llemma}        & $4.77$ \\
arXiv \citep{azerbayev2024llemma}                & $0.88$ \\
ShareGPT \citep{sharegpt_vicuna_unfiltered}      & $0.21$ \\
Gutenberg books \citep{manu2023gutenberg}        & $0.07$ \\
\bottomrule
\end{tabular}
\caption{Training mixture, as a share of documents.}
\label{tab:apx_mixture}
\end{table}

Documents are embedded with \citet{nussbaum2024nomic} and partitioned by $k$-means into $1200$ clusters ($1150$ for training, $50$ held out for validation). Each training step draws from $32$ clusters; within a step, a sequence is built from documents of a single cluster, delimited by \texttt{[bos]} and \texttt{[eos]}, concatenated and truncated at $l$.

\subsection{Evaluation clusters}
\label{apx:eval_clusters}

The same clustering procedure gives $50$ clusters on Simple English Wikipedia. The corpus contains large families of near-template stubs --- United States counties, Italian \emph{comuni}, French communes --- and $k$-means naturally collects each family into a cluster of its own:

\begin{quote}\small
\emph{Saint-Cr\'eac is a commune in the Gers department. It is in southwestern France. Related pages: Communes of the Gers department.}

\emph{Saint-Georges is a commune in the Gers department. It is in southwestern France. Related pages: Communes of the Gers department.}

\emph{Tourdun is a commune in the Gers department. It is in southwestern France. Related pages: Communes of the Gers department.}
\end{quote}

\begin{figure}[H]
    \centering
    \includegraphics[width=0.3\linewidth]{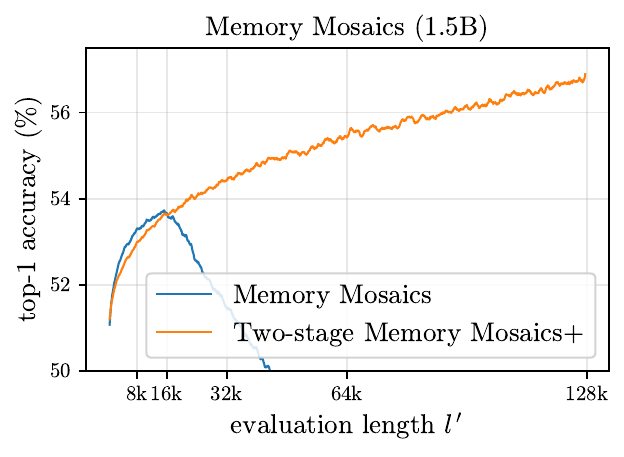}
    \caption{Memory Mosaics and two-stage Memory Mosaics$+$ on all $50$ Simple English Wikipedia
    clusters, at $1.5$B. As Figure~\ref{fig:two-stage_vs_one-stage}, without the cluster selection.}
    \label{fig:all_clusters}
\end{figure}

We remove these clusters, identified from the documents alone and not from any model's predictions, to avoid trivial solutions. That leaves $19$ clusters, used in all main-text SimpleWiki evaluations. Figure~\ref{fig:all_clusters} repeats the comparison on all $50$: the conclusion is unchanged.

\section{Evaluation on Simple English Wikipedia without clustering}
\label{apx:simplewiki_no_cluster}

The whole of Simple English Wikipedia is treated as one pool of related tasks, with no clustering. Only the evaluation changes. The training and the evaluation sequence construction are the same as in Figure~\ref{fig:two-stage_vs_one-stage}. The pattern is unchanged --- two-stage models keep improving, one-stage models peak and decline --- so the conclusion does not rest on the clustering step. The two figures cover different portions of the evaluation corpus, so their shapes are comparable and their magnitudes are not.
\begin{figure}[H]
\vspace{-0.4em}
    \centering
    \includegraphics[width=0.235\linewidth]{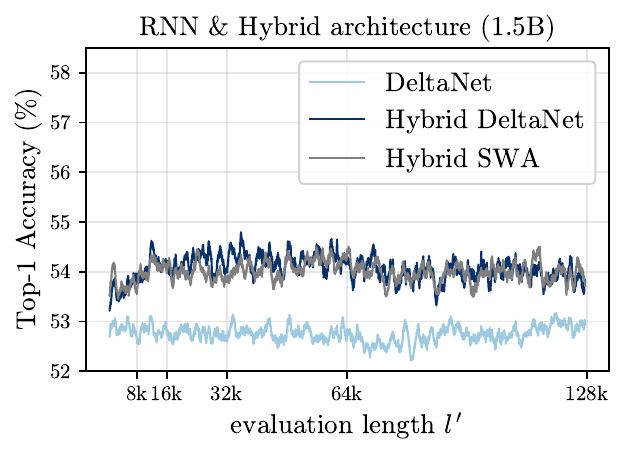}
    \includegraphics[width=0.235\linewidth]{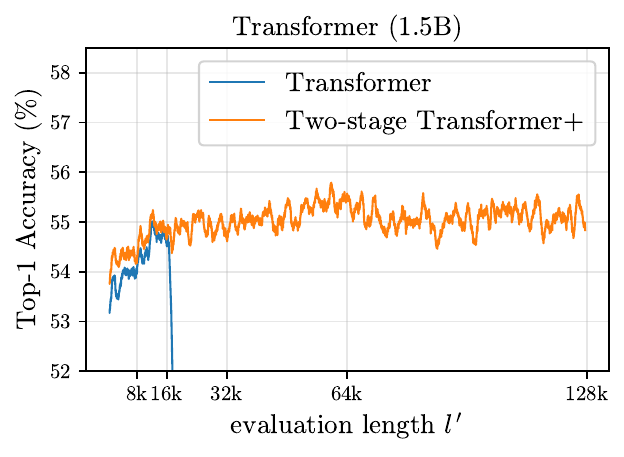}
    \includegraphics[width=0.235\linewidth]{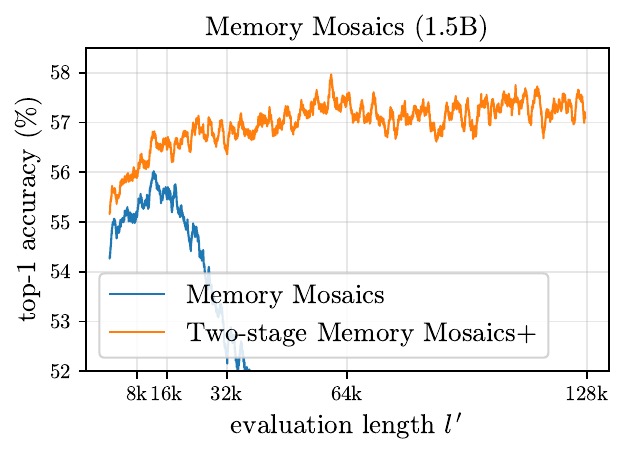}
    \caption{Accuracy as the number of related tasks in the context grows, at $1.5$B. As
    Figure~\ref{fig:two-stage_vs_one-stage}, but the whole of Simple English Wikipedia is treated
    as related tasks, with no clustering.
    \textbf{1)} RNN and hybrid architectures; \textbf{2)} Transformer versus two-stage Transformer$+$; \textbf{3)} Memory Mosaics versus two-stage Memory Mosaics$+$.}
    \label{fig:no_cluster_two-stage_vs_one-stage}
    \vspace{-1.3em}
\end{figure}

\section{Memory Mosaics with matched parameters}
  \label{apx:param_matched}

Memory Mosaics has no query projection, so with the same head configuration its learning module has 25\% fewer
parameters than in the other architectures\footref{fn:params}. Table~\ref{tab:param_matched}
closes that gap at the 1.5B scale by using 43 heads of dimension 64 instead of 16 heads of dimension 128. The
key, value, and output projections then have $3 \times 2048 \times (43 \cdot 64) \approx 16.9$M parameters
  per block, against $4 \times 2048^2 \approx 16.8$M for attention in the other architectures, a match within
  1\%. Data, optimizer, and the fine-tuning protocol are unchanged. The table shows that a small engineering
  change shifts the accuracy level while leaving the sign of $\Delta$ intact, which is why the main text makes
  qualitative comparisons rather than quantitative ones.
  
  \begin{table}[H]
  \centering
  \small
  \begin{tabular}{@{}l cc c@{}}
  \toprule
   & \multicolumn{2}{c}{Training length $l$} & \\
  \cmidrule(lr){2-3}
  Model & 4k & 64k & $\Delta>0$ \\
  \midrule
  Memory Mosaics ($16~\text{heads} \times 128~\text{dim}$) & 25.93 & 28.58 & \textcolor{blue}{\ding{52}} \\
  Memory Mosaics ($43~\text{heads} \times 64~\text{dim}$) & 28.73 & 30.78 & \textcolor{blue}{\ding{52}} \\
  \bottomrule
  \end{tabular}
  \caption{\textsc{gsm8k} accuracy (\%) after fine-tuning as in Figure~\ref{fig:rich_sft}, at 1.5B. $\Delta$ is (64k $-$ 4k).}
  \label{tab:param_matched}
  \end{table}

\section{Position encoding on the long-range attention}
\label{apx:two-stage-transformer+_ablation}

\begin{figure}[H]
    \centering
    \includegraphics[width=0.3\linewidth]{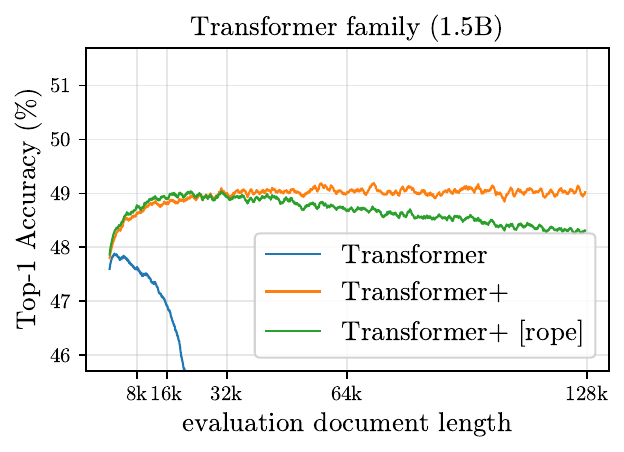}
    \caption{As Figure~\ref{fig:scale_l}, at $1.5$B, with a third curve: Transformer$+$ whose long-range attention also carries \textsc{rope}.}
    \label{fig:transformer_rope}
\end{figure}

Transformer$+$ applies \textsc{rope} to the short sliding window and none to the long-range attention above it. Figure~\ref{fig:transformer_rope} repeats the setting of Figure~\ref{fig:scale_l} with a third curve, a Transformer$+$ whose long-range attention also carries \textsc{rope}. Transformer$+$ with \textsc{rope} is worse than Transformer$+$, confirming the choice of using \textsc{nope} in the long-range attention, as in the main text.

\section{Sliding-window length in Hybrid \textsc{swa}}
\label{apx:swa_length}

  \begin{table}[H]
  \centering
  \small
  \setlength{\tabcolsep}{4pt}
  \resizebox{0.5\textwidth}{!}{
  \begin{tabular}{@{}l  cccc  cccc@{}}
  \toprule
  ~~~~~~~~~$l^\prime$
  & \multicolumn{4}{c}{\emph{Hybrid \textsc{swa}-4k}}
  & \multicolumn{4}{c}{\emph{Hybrid \textsc{swa}-2k}} \\
  \cmidrule(lr){2-5}\cmidrule(lr){6-9}
  $l$
  & 4k & 8k & 16k & 32k
  & 4k & 8k & 16k & 32k \\
  \midrule
  4k  & \cellcolor{diagred}47.94 & --- & --- & ---
      & \cellcolor{diagred}48.19 & --- & --- & --- \\
  8k  & 47.97 & \cellcolor{diagred}47.90 & --- & ---
      & 47.88 & \cellcolor{diagred}47.80 & --- & --- \\
  16k & 47.91 & 47.91 & \cellcolor{diagred}48.00 & ---
      & 47.97 & 47.98 & \cellcolor{diagred}48.01 & --- \\
  32k & 47.68 & 47.69 & 47.71 & \cellcolor{diagred}47.62
      & 47.65 & 47.60 & 47.65 & \cellcolor{diagred}47.53 \\
  64k & \cellcolor{longblue}47.54 & \cellcolor{longblue}47.54 & \cellcolor{longblue}47.64 & \cellcolor{longblue}47.49
      & \cellcolor{longblue}47.54 & \cellcolor{longblue}47.55 & \cellcolor{longblue}47.60 & \cellcolor{longblue}47.46 \\
  \midrule
  $\Delta>0$
      & \textcolor{red}{\ding{56}} & \textcolor{red}{\ding{56}} & \textcolor{red}{\ding{56}} &
  \textcolor{red}{\ding{56}}
      & \textcolor{red}{\ding{56}} & \textcolor{red}{\ding{56}} & \textcolor{red}{\ding{56}} &
  \textcolor{red}{\ding{56}} \\
  \bottomrule
  \end{tabular}}
  \caption{Top-1 accuracy (\%) on SimpleWiki for Hybrid \textsc{swa} with a sliding-window
  length of 4k vs 2k, under the same protocol as Table~\ref{tab:rich}. The window length does not
  change the conclusion: for both, a longer max training length $l$ does not help the model learn
  new tasks.}
  \label{tab:swa_length_ablation}
  \end{table}

\section{What implements the short memory}
\label{apx:ablation_short_mem}
\begin{figure}[H]
    \centering
    \includegraphics[width=0.3\linewidth]{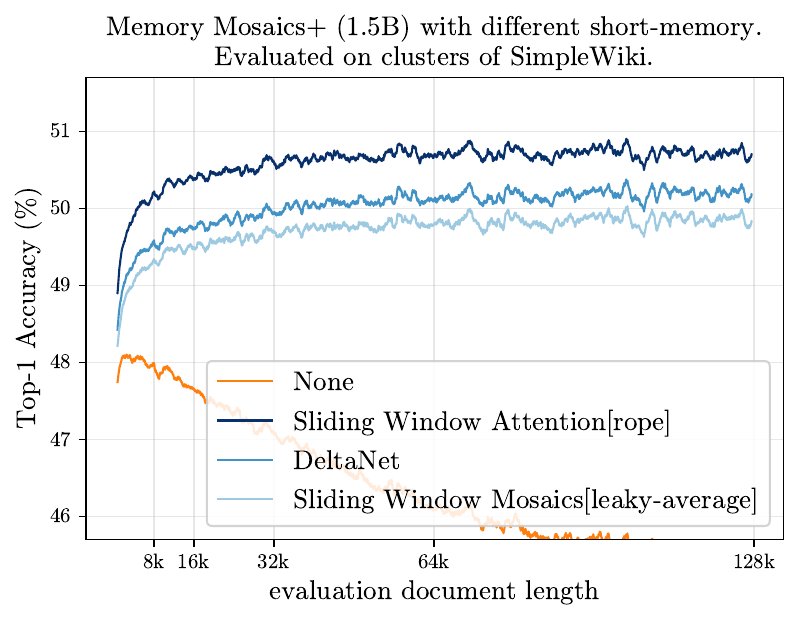}
    \caption{As Figure~\ref{fig:scale_l}, at $1.5$B, sweeping the short-memory implementation in Memory Mosaics$+$: attention with \textsc{rope}, a Mosaics module with the leaky-average key extractor, and DeltaNet.}
    \label{fig:mm_shortmem}
\end{figure}

Section \ref{sec:exp:inference} argues for a short-memory component to handle highly dependent local tokens, but says nothing about its type. Figure~\ref{fig:mm_shortmem} sweeps that short memory in Memory Mosaics$+$ over attention with \textsc{rope}, a Mosaics module with the leaky-average key extractor, and DeltaNet, in the setting of Figure~\ref{fig:scale_l}. The curve ``None'' indicates no short memory, i.e. Memory Mosaics.
The other three curves follow the same trend: what matters is a separate short memory, not how it is implemented.

\section{Per-head gate and position encoding}
\label{apx:mm_mem_gate}

Table~\ref{tab:head_ablation} repeats the comparison of Table~\ref{tab:heads} with two ablations: Memory Mosaics without the per-head sigmoid gate, and a Transformer without any position encoding [\textsc{nope}]. Neither changes the sign. Without the gate, Memory Mosaics still gains as $G$ grows, so the gate is not what produces the gain. Without \textsc{rope}, the Transformer still does not gain, so position encoding is not what prevents it; the same holds for Hybrid \textsc{swa}, whose full-attention layers already carry no position encoding. Transformer [\textsc{nope}] is lower overall, as expected: with no position encoding anywhere, it cannot model the dependence between nearby tokens. Only the sign matters here.

\begin{table}[H]
      \centering
      \resizebox{0.7\linewidth}{!}{
  \begin{tabular}{@{}lccccc c@{}}
  \toprule
   & \multicolumn{5}{c}{Number of groups $G$ (heads)} & \\
  \cmidrule(lr){2-6}
  Model & 4 & 8 & 16 & 32 & 64 & $\Delta>0$ \\
  \midrule
  Memory Mosaics      & \cellcolor{diagred}48.43 & 48.63 & 48.97 & \textbf{49.07} & \cellcolor{longblue}49.02 &
  \textcolor{blue}{\ding{52}} \\
  Memory Mosaics [no per-head gate] & \cellcolor{diagred}48.15 & 48.29 & 48.46 & \textbf{48.52} & \cellcolor{longblue}48.47 &
  \textcolor{blue}{\ding{52}} \\
  Transformer  & \cellcolor{diagred}48.31 & \textbf{48.38} & 48.05 & 47.92 & \cellcolor{longblue}47.08 & \textcolor{red}{\ding{56}}
  \\
  Transformer [\textsc{nope}] & \cellcolor{diagred}45.27 & \textbf{45.44} & 45.36 & 45.37 & \cellcolor{longblue}44.40 &
  \textcolor{red}{\ding{56}} \\
  Hybrid \textsc{swa}   & \cellcolor{diagred}\textbf{48.00} & 47.95 & 47.97 & 47.67 & \cellcolor{longblue}46.85 &
  \textcolor{red}{\ding{56}} \\
  \bottomrule
  \end{tabular}}
  \caption{Top-1 accuracy (\%) on SimpleWiki, evaluated at length $l'=4$k; all models trained at length $l=16$k. As
  Table~\ref{tab:heads}, with two ablations added: Memory Mosaics without the per-head gate, and Transformer without position
  encoding [\textsc{nope}]. Model width is fixed, so raising the number of heads raises the number of groups $G$ and shrinks each
  group. Row maxima in bold. $\Delta$ is (\textcolor{blue}{64 heads} $-$ \textcolor{red}{4 heads}).}
      \label{tab:head_ablation}
  \end{table}
  
\section{The implicit balancing weight}
\label{apx:two-stage_magnitudes}

\begin{figure}[H]
    \centering
    \includegraphics[width=0.3\linewidth]{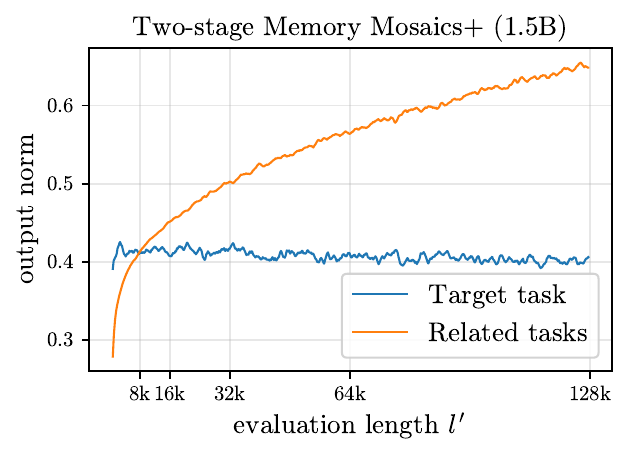}
    \caption{Output magnitude of the target-task and related-task parts of the long memory in two-stage
    Memory Mosaics$+$ at $1.5$B, against evaluation length.}
    \label{fig:two-stage_memory_norm}
\end{figure}

The two-stage models never compute the optimal weight $\alpha^{*}=(\mathrm{Var}_{r}+\delta^{2})/(\mathrm{Var}_{t}+\mathrm{Var}_{r}+\delta^{2})$ of Section~\ref{sec:two_predictions}; each part is normalized on its
own, so the two contributions enter in proportion to their output magnitudes. That magnitude is informative because each part is a normalized average of unit-norm values: its norm
approaches $1$ when the retrieved values agree and falls towards $0$ when they disagree. Figure~\ref{fig:two-stage_memory_norm} plots the two against evaluation length. The target-task
part stays flat while the related-task part grows --- the related evidence is becoming more concordant as it accumulates --- so the balance shifts towards it with nothing computing
a weight. This reproduces the direction in which $\alpha^{*}$ moves as related tasks accumulate, though not its value; estimating the weight directly is left to future work.

\end{document}